%% file: acl_latex.tex
\documentclass[11pt]{article}

\usepackage[final]{acl}
\usepackage{times}
\usepackage{latexsym}
\usepackage[T1]{fontenc}

\usepackage[utf8]{inputenc}

\usepackage{microtype}

\usepackage{inconsolata}

\usepackage{graphicx}
\usepackage{array}
\usepackage{mathtools}
\usepackage{colortbl}   
\usepackage{multirow}
\usepackage{amsmath,amssymb}
\usepackage{bbm}

\usepackage{tabularx}
\usepackage{tcolorbox}
\usepackage{makecell}
\usepackage{adjustbox}
\usepackage{booktabs}
\usepackage{soul}

\title{Data-Driven Personas for Survey Simulation: Insights into Simulation Alignment Across Data-Access Regimes}

\author{%
  \parbox{0.45\textwidth}{%
    \centering
    \textbf{Dongryeol Lee}\textsuperscript{$\ast$} \\
    \normalfont Seoul National University \\
    \texttt{drl123@snu.ac.kr}
  }%
  \parbox{0.45\textwidth}{%
    \centering
    \textbf{Weronika Łajewska} \\
    \normalfont Amazon \\ 
    \texttt{lajewska@amazon.com}
  }\\[3em]%
  \parbox{0.45\textwidth}{%
    \centering
    \textbf{Leonardo Perelli} \\
    \normalfont Amazon \\ 
    \texttt{lperelli@amazon.com}
  }%
  \parbox{0.45\textwidth}{%
    \centering
    \textbf{Saab Mansour} \\
    \normalfont Amazon \\
    \texttt{saabm@amazon.com}
  }%
}

\newcommand{\wl}[1]{\textcolor{orange}{#1}}

\usepackage{enumitem}

\begin{document}
\maketitle
\begingroup
\renewcommand{\thefootnote}{}
\footnotetext{\textsuperscript{$\ast$} Work done as an intern at Amazon.}
\endgroup

\setcounter{footnote}{0}

\input{0.abstract}
\input{1.intro}
\input{2.rel_work}
\input{3.formulation}
\input{4.setup}

\input{5.experiments}
\input{6.discussion}
\input{7.conclusion}
\input{limitation}

\input{ethic}
\bibliography{custom}
\newpage
\appendix

\input{appendix}
\end{document}

%% file: 0.abstract.tex
\begin{abstract}
Large language models (LLMs) offer new opportunities for public opinion research by enabling early prediction of survey responses, potentially reducing the cost and time of traditional surveys. 
However, many existing steering approaches rely on target-domain human data for fine-tuning or prompting that is costly to collect and raises privacy concerns. 
In this paper, we study demographic group-level survey simulation, where personas induced from heterogeneous, anonymized public behavioral data condition agents that simulate responses of individuals from specific demographic groups.
We examine whether representative personas can be induced from diverse sources and analyze how the domain, scale, and granularity of the source data affect survey simulation alignment.
We find that personas induced from out-of-domain sources rarely outperform simulations conditioned only on basic demographic information, largely due to population mismatch. 
However, when personas are accurately assigned to the target demographic groups, alignment improves substantially. 
Finally, personas induced from target-domain survey data generalize better as more survey question history becomes available, suggesting that richer behavioral evidence enables more stable persona trait inference that transfers to better unseen questions simulation alignment.
\end{abstract}

%% file: 1.intro.tex
\section{Introduction}
Surveys are an essential tool for understanding public opinion and informing decisions in public policy, industry, and social science research. Because attitudes and preferences evolve over time and vary across demographic groups, organizations often rely on repeated and large-scale surveys to capture these differences. However, collecting survey data is costly, time-consuming, and logistically demanding.
Recent advances in large language models (LLMs) have enabled their use for \emph{survey simulation}, where models are prompted to predict the opinions of different demographic groups~\cite{wang2025user, gao2024large, manning2024automated, bail2024can}.
This approach offers a scalable alternative to traditional user studies while enabling faster exploration of public opinion dynamics~\cite{argyle2023out, aher2023using, hewitt2024predicting}.
\input{figures/figure1}

Prior work on survey simulation has primarily explored prompt engineering~\cite{santurkar2023whose, kim2023ai}, persona-based conditioning~\cite{park2024generative}, and fine-tuning~\cite{cao2025specializing, kolluri2025finetuning, suh2025language} to improve the alignment between simulated and human responses.
However, much of this work relies on target-domain human data for learning and control---for example, by using interviews or other individual-level records in prompts
~\cite{park2024generative}, or fine-tuning models on data drawn from the same source as the evaluation set~\cite{cao2025specializing, kolluri2025finetuning, suh2025language}.
Collecting such data is resource-intensive and often constrained
by privacy regulations, consent requirements, and governance frameworks, while also providing limited coverage of underrepresented subpopulations.

In this work, we investigate the impact of synthetic personas induced from heterogeneous sources of human behavioral data on steering LLM-based survey simulations.~\footnote{All experiments use publicly available datasets with anonymized behavioral traces, and no personally identifiable information was used or collected in this research. 
To support reproducibility, we release the code at \url{https://github.com/amazon-science/data-driven-personas}, but not raw user-level traces, record-linked personas, or intermediate artifacts that could increase re-identification risk.}
Our goal is to understand when such personas, defined as natural-language profiles distilled from behavioral traces, improve the alignment of survey simulation.
This question is important in practice because practitioners often have uneven access to persona source data.
In some settings, only external behavioral data are accessible (\textbf{Low access / out-of-domain}); in others, additional information such as ground-truth (GT) demographics of participants may be accessible (\textbf{Mid access / out-of-domain with GT demographics}); and in the most favorable cases, 
part of the target survey data itself may be available
(\textbf{High access / in-domain}). 
To reflect these realistic constraints, we frame 
persona-conditioned survey simulation in terms of \textbf{data-access regimes} and study how the effectiveness of personas changes as stronger forms of supervision become available.

Accordingly, we treat persona induction from behavioral data and downstream survey simulation as separate tasks. 
We first induce personas from different sources of behavioral data, then use them to condition individual agents, and evaluate the resulting aggregated predictions against human response distributions at the demographic-group level (e.g., the Democrat group distribution
in Figure~\ref{figure1}).~\footnote{We do not aim to replicate individual behavior, but to statistically approximate group-level responses.}

Across survey benchmarks, we find that synthetic personas are not uniformly beneficial for simulation---their effectiveness depends strongly on the available data and the degree of source--target alignment. 
Nevertheless, the value of persona-conditioned simulation should not be judged solely by aggregate alignment. 
Persona-based simulation can also offer an important advantage in interpretability: each simulated response is linked to a specific persona induced from behavioral evidence, and aggregating over multiple personas allows the simulator to represent within-group heterogeneity rather than collapsing a group into a single demographic descriptor. 
We therefore view personas not only as a potential means of improving aggregate alignment, but also as a way to make survey simulation more behaviorally grounded, more interpretable, and more sensitive to within-group variation. 
More broadly, our results provide practical guidance for deploying survey simulation under realistic data constraints, showing that effective simulation depends not only on the richness of persona information, but also on the type of behavioral data, accuracy of persona to group assignment, and source--target population mismatch (i.e., the degree to which the source behavioral population resembles the target survey respondents).

Our contributions are summarized as follows:
\begin{itemize}[left=0.2em,itemsep=0em]
    \item We show that personas induced from out-of-domain data do not generalize well to survey simulation, 
    largely because demographic attribute misassignment causes personas to be mapped to inappropriate target groups.
    \item We show that personas induced from in-domain data generalize better to held-out questions as more history is provided. 
    They also yield the strongest and most stable alignment across different sizes of demographic groups, with especially large gains for smaller, underrepresented groups, and perform best across all demographic attributes.

\end{itemize}


%% file: figures/figure1.tex
\begin{figure}[t]
\centering
\includegraphics[width= 0.95\columnwidth]{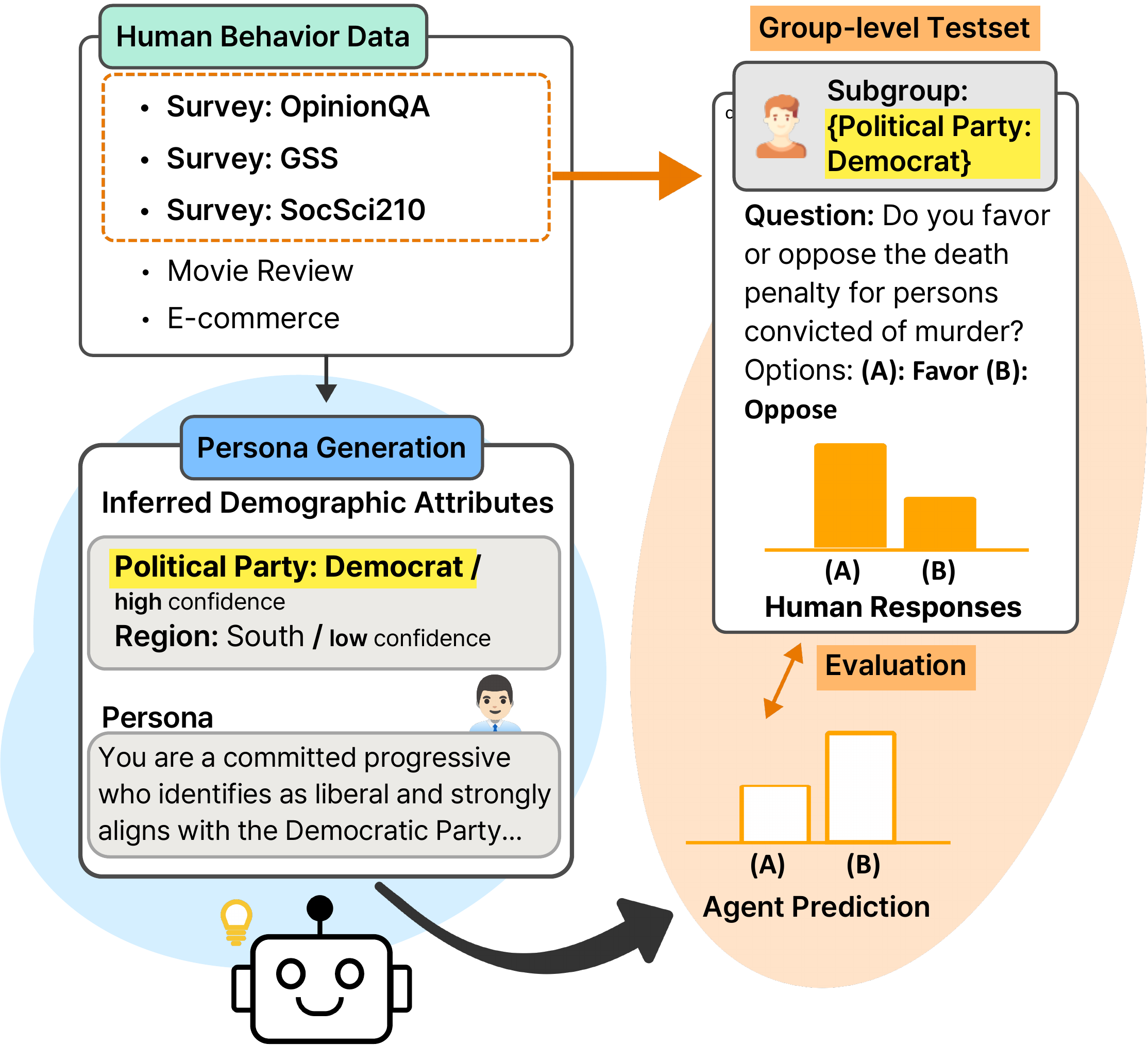} 
\caption{Data driven persona induction and survey simulation: In our framework, we generate personas from behavioral data across different domains and predict responses for individuals in target demographic groups.}
\label{figure1}
\end{figure}

%% file: 2.rel_work.tex
\section{Related Work}
\paragraph{Survey simulation.}
Recent work explores LLMs as a lower-cost alternative to large-scale human data collection, demonstrating their ability to reproduce some group-level response patterns in surveys and behavioral, economic, and political settings~\cite{argyle2023out, aher2023using, horton2023large, hewitt2024predicting}.
However, subsequent studies raised concerns about validity, robustness, and representational bias~\cite{dominguez2024questioning, bisbee2024synthetic}.
Subsequent efforts proposed prompting and elicitation methods, including sociodemographic prompting, persona-based conditioning, and elicitation of response distributions~\cite{santurkar2023whose, moon2024virtual, park2024generative, manning2024automated, kwok2024evaluating, sun2024random}.
Yet capturing within-group heterogeneity and avoiding stereotypical predictions remain open challenges~\cite{park2024diminished, kapania2025simulacrum, cheng2023compost}, and fine-tuning approaches rely on survey or experimental response data for adaptation~\cite{cao2025specializing, kolluri2025finetuning, suh2025language}.
In contrast, we focus on simulation of \emph{group-level response distributions} and examine how conditioning signals transfer across domains and data-access regimes.

\paragraph{Synthetic personas.}
Synthetic persona generation constructs artificial user profiles to support downstream tasks such as political and economic simulation~\cite{tseng2024two, chen2024persona, horton2023large}.
Earlier approaches often relied on survey or census data~\cite{argyle2023out, chang2025llms}, demographic profiles~\cite{hwang2023aligning, santurkar2023whose}, or political identities~\cite{simmons2023srw} to steer model behavior.
With LLMs, recent work has moved toward descriptive personas derived from richer sources, including web text~\cite{ge2024scaling}, interview transcripts~\cite{park2024generative}, personal narratives~\cite{moon2024virtual}, and shopping histories~\cite{mansour2025paars}, with behavioral-data-grounded personas also supporting A/B test simulation~\cite{benomar2026data}.
Other work further explores automatically refined personalization: few-shot personalization~\cite{kim2025few} and collaborative data refinement~\cite{sun2025persona}.
These methods provide richer conditioning signals, but approaches grounded in individual records may depend on costly or sensitive target-domain data, and demographic prompting can yield caricatured simulations~\cite{cheng2023compost}.
In contrast, we focus on \emph{data-driven personas induced from external behavioral sources} to steer \emph{distribution-level} survey simulation, enabling out-of-domain deployment without individual-level target-survey records.

%% file: 3.formulation.tex
\section{Survey Simulation under Different Data-access Regimes}
\label{sec3}

In this work, we study \emph{agentic survey simulation}, in which a pool of LLM-based agents predict how humans would respond to survey questions~\cite{argyle2023out, wang2025user}. 
Our central question is whether---and under what conditions---\emph{synthetic personas}, defined as natural-language profiles inferred from behavioral data, improve simulation alignment for survey responses. 
In our setting, agents are conditioned on synthetic personas, while evaluation is performed on aggregated response distributions for each demographic group~\cite{suh2025language, cao2025specializing, kolluri2025finetuning}.


\input{figures/figure2}

\subsection{Persona-conditioned Simulation}
\label{sec_3.1}

Let $Q$ be a set of survey questions. 
Each question $q \in Q$ has answer options $X_q = \{x_1, \ldots, x_{N_q}\}$. 
For a demographic group $g$, defined by a single attribute--value pair (e.g., \texttt{AGE=30--49}), empirical human responses induce a subgroup distribution $U_{q,g}$ over $X_q$. 
Our goal is to produce a simulated distribution $S_{q,g}$ that is closely aligned with
$U_{q,g}$. 

To improve $S_{q,g}$, we condition each agent on a synthetic persona. 
Our framework proceeds in two stages: 1) persona generation and 2) persona-conditioned survey simulation. 
In the first stage, we induce a pool of personas from behavioral data. 
Each persona is a natural-language description of the individual's characteristics and is associated with inferred demographic attributes accompanied by corresponding confidence scores (see Figure~\ref{figure2}), enabling it to be matched to a target demographic group $g$. 
In the second stage, these induced personas condition individual agents to simulate responses for a target survey, and the resulting predictions are aggregated for each demographic group.

For a target question $q$ and group $g$, we select personas that match the demographic attribute defining $g$, prompt $N_A$ LLM agents---one per persona---to produce a \textit{verbalized distribution}~\cite{hu2025simbench} over the answer options, and then aggregate outputs:~\footnote{Based on the findings of our pilot study (Appendix~\ref{app:pilot}), we set $N_A = 30$, select the highest-confidence matching personas for each demographic group, and use verbalized distributions to represent agent outputs.}
$S_{q,g}(x) = \frac{1}{N_A} \sum_{j=1}^{N_A} S^{(j)}_{q,g}(x), \quad x \in X_q$.
This formulation separates persona construction from downstream evaluation, allowing the same simulation protocol to be instantiated with different persona sources and different levels of access to target-domain information.

\subsection{Data Access Regimes}
\label{sec_3.2}

We organize our experiments based on the information available for persona construction and for assigning personas to demographic groups. Each data-access regime motivates a corresponding research question.

\paragraph{Low-access / out-of-domain.}
This regime represents situations where practitioners have no access to target-survey data. 
Personas must instead be induced from external sources, such as non-survey behavioral data, survey data from other datasets or waves, or the LLM’s parametric knowledge. It mimics scenarios where only general behavioral evidence is available, testing whether such out-of-domain personas can still produce meaningful predictions for a target survey. This motivates
\emph{RQ1: Can personas induced from out-of-domain behavioral data improve survey simulation alignment?}

\paragraph{Mid-access / out-of-domain setting with GT demographics}
This regime reflects settings where practitioners have behavioral data from external sources along with GT demographic information. Personas can be induced using both behavioral traces and demographic attributes. This scenario captures cases where demographic alignment is feasible even without direct access to the target survey, allowing us to study how correct persona–group assignment affects simulation alignment. This motivates \emph{RQ2: How does access to ground-truth demographic information in out-of-domain behavioral data improve survey simulation alignment?}


\paragraph{High-access / in-domain.}
This regime represents scenarios where practitioners have partial access to the target survey itself. 
Personas are induced from these in-domain responses and used to predict held-out questions within the same survey. 
It captures the potential of leveraging partial in-domain data to improve simulation for unseen questions, motivating \emph{RQ3: Can in-domain personas induced from partial survey responses generalize to predict held-out questions?}

%% file: figures/figure2.tex
\begin{figure}[t]
\centering
\includegraphics[width= 0.95\columnwidth]{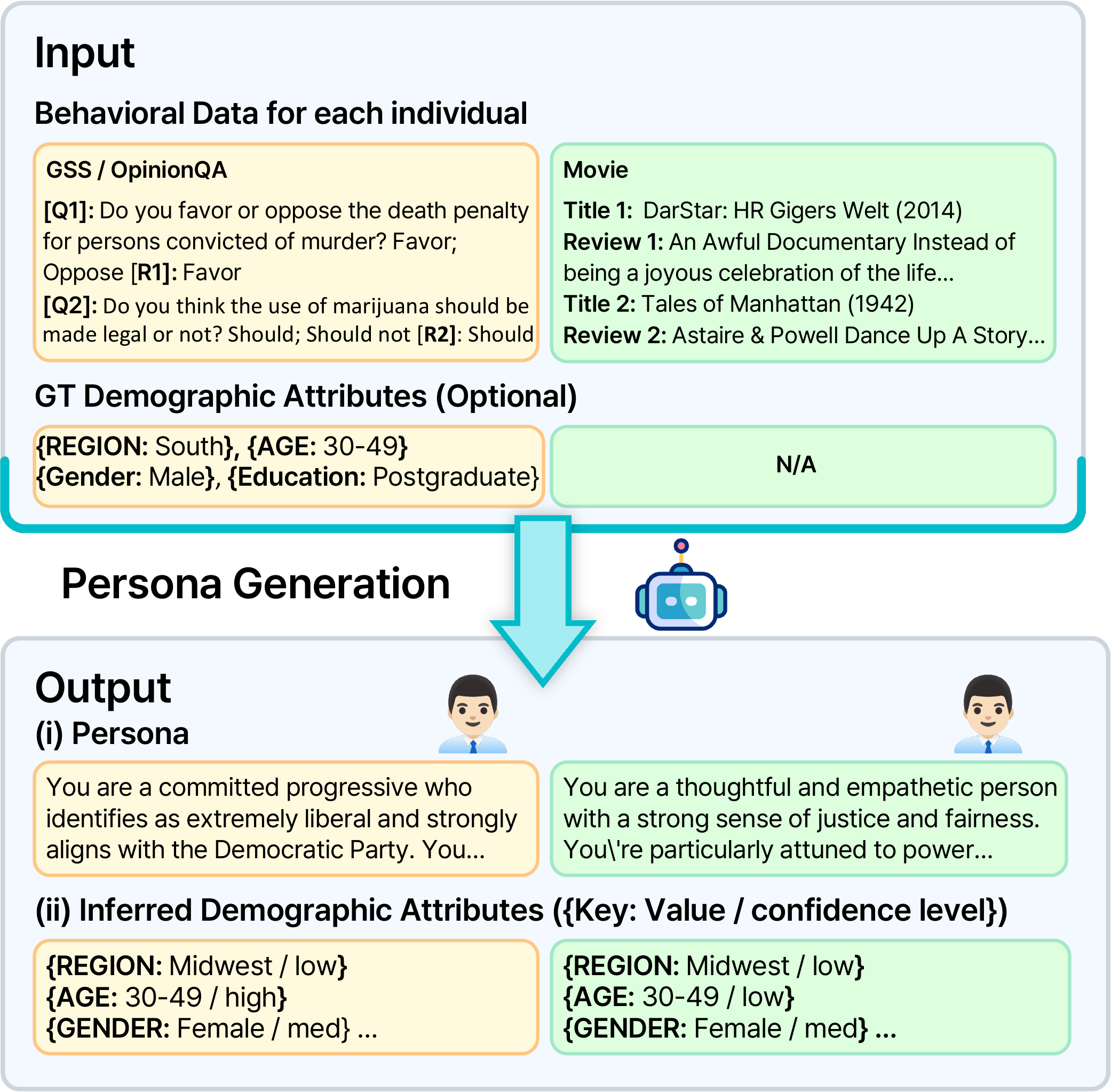} 
\caption{Persona generation examples across domains.}
\label{figure2}
\vspace{-4mm}
\end{figure}

%% file: 4.setup.tex
\section{Experimental Setup}
\label{sec_4}
We now instantiate the study formulation described in Section~\ref{sec3} with concrete target datasets, persona sources, baselines, and evaluation metrics.

\subsection{Target Survey Datasets Construction}

We evaluate survey simulation on five publicly available survey datasets or waves: OpinionQA Waves 26 and 29~\cite{santurkar2023whose}, the General Social Survey (GSS)~\cite{davern2024gss}, and two SocSci210 subsets, MP3DR and NJ5DX~\cite{kolluri2025finetuning}.~\footnote{OpinionQA waves and SocSci210 subsets differ in topics and populations and are therefore treated as separate surveys.}
Each dataset provides respondent-level demographic attributes and individual survey responses, enabling construction of group-level empirical response distributions.

Following Section~\ref{sec3}, we  construct demographic group-level test instances by aggregating responses from respondents who share a given demographic attribute value. 
This yields test instances of the form $(q_i, \{x_k\}, g, U_{i,g})$.~\footnote{Appendix~\ref{app:dataset} covers statistics of both target dataset and persona-source datasets and shows example of test instance.}

\subsection{Persona Sources}

We construct personas using three types of data.

\paragraph{Model knowledge.}
Given a target attribute defining the demographic group $g$, we prompt Claude 3.7 Sonnet~\cite{claude} to generate a diverse set of personas consistent with that group. This provides a source-free reference point that relies only on the model's parametric knowledge.

\paragraph{Non-survey behavioral data.}
We induce personas from anonymized movie review histories~\cite{yoon2024evaluating} and e-commerce shopping histories~\cite{berke2024open}. 
This instantiates the low-access setting where only external out-of-domain behavioral traces are available.

\paragraph{Survey behavioral data.}
Personas are generated from anonymized respondent-level survey responses. When the source survey differs from the target, this represents an out-of-domain setting. When the source survey is the same as the target survey, we split the questions into persona-induction and evaluation subsets, using the former as the behavioral data source and the latter as held-out test questions, representing the in-domain setting.

\subsection{Persona Generation and Selection}
\label{sec_4.3}
For data-driven persona generation, we use Claude 3.7 Sonnet~\cite{claude} as the persona generator. As shown in Figure~\ref{figure2}, given an individual's behavioral record, the model produces: (i) a natural-language persona for simulation, and (ii) inferred demographic attributes for 11 keys (CREGION, AGE, SEX, EDUCATION, MARITAL, RELIG, RELIGATTEND, POLPARTY, POLIDEOLOGY, INCOME, and RACE) with confidence scores. Each key has a fixed closed set of options (e.g., AGE: 18--29 | 30--49 | 50--64 | 65+; POLPARTY: Democrat | Republican | Independent | Other); full option sets are listed in Appendix~\ref{app:persona}.

For each demographic group $g$, we assign personas using their inferred demographic attributes, then rank candidates by confidence score and select the top $N_A = 30$ per group for simulation. This confidence-based selection is motivated by our pilot study (Appendix~\ref{app:pilot}), which shows that high-confidence extraction yields substantially higher demographic accuracy (e.g., 0.81 vs.\ 0.39 on GSS; Table~\ref{tab:conf_acc}), and that simulation performance largely saturates beyond 30 personas.

For settings where ground-truth demographic labels are available for source individuals \emph{(Mid access / out-of-domain with GT demographics)}, we consider two additional variants. 
In \textit{GT assignment}, ground-truth demographics are used only for the assignment of personas to different demographic groups.
In \textit{GT generation}, ground-truth demographic information is also provided as an additional data source during persona induction.

\subsection{Baselines and Simulation Protocol}

We compare the persona-conditioned simulation against three baselines following SimBench~\cite{hu2025simbench}.~\footnote{All baselines use the same individual-agent simulation and group-level aggregation procedure.}

\paragraph{B1 (No demographic context).}
Each agent receives only the question and answer options.

\paragraph{B2 (Country-level context).}
Each agent receives a short demographic context (e.g. \emph{You are from USA}) in addition to the question.

\paragraph{B3 (Single-attribute context).}
Each agent receives an explicit value of a  demographic attribute (e.g. \emph{You are Male}) along with the question.

\paragraph{Simulation protocol.}
For each target question $q$ and demographic group $g$, we prompt $N_A = 30$ LLM agents in parallel, each conditioned on a distinct persona selected as described in Section~\ref{sec_4.3}. 
Each agent produces a verbalized probability distribution over the answer options (e.g., \texttt{A: 30\%, B: 40\%, C: 30\%}), and the  resulting distributions are aggregated via simple averaging as described in Section~\ref{sec3}. 
This aggregation approximates within-group heterogeneity by pooling over a diverse set of persona-conditioned agents rather than  collapsing each group to a single representative response. 

\input{tables/table_main}

\subsection{Evaluation Metrics}
We quantify the misalignment between the human demographic group-level response distribution $U_{i,g}(x)$ and the simulated response distribution $S_{i,g}(x)$ using two complementary metrics: Total Variation Distance (TVD)~\cite{hu2025simbench, lajewska2026radius} and 1-Wasserstein distance (WD)~\cite{suh2025language,moon2024virtual} (see Appendix~\ref{app:metric} for details). 
%
For each test instance $(q_i,g)$, we compute the distance between $U_{i,g}$ and $S_{i,g}$ and report the average across all questions and demographic groups. Lower values indicate closer alignment for both metrics.




%% file: tables/table_main.tex
\begin{table*}[t]
\centering
\scriptsize
\renewcommand{\arraystretch}{1.1}
\small
\setlength{\tabcolsep}{4pt}
\begin{tabular}{lcccccccccc}
\toprule

\multirow{2}{*}{Baseline/Persona Source}
& \multicolumn{2}{c}{GSS}
& \multicolumn{2}{c}{OpinionQA W26}
& \multicolumn{2}{c}{OpinionQA W29}
& \multicolumn{2}{c}{SocSci210 (MP3DR)}
& \multicolumn{2}{c}{SocSci210 (NJ5DX)} \\
\cmidrule(lr){2-3} \cmidrule(lr){4-5} \cmidrule(lr){6-7} \cmidrule(lr){8-9} \cmidrule(lr){10-11}
& TVD & WD & TVD & WD & TVD & WD & TVD & WD & TVD & WD \\
\midrule

B1 & 0.169 & 0.322 & 0.189 & 0.299 & 0.171 & 0.262 & \underline{0.311} & \textbf{0.474} & 0.241 & 0.678 \\
B2 & \underline{0.160} & \underline{0.301} & \textbf{0.173} & \textbf{0.274} & \underline{0.139} & \underline{0.204} & 0.314 & 0.552 & 0.204 & 0.560 \\
B3 & \textbf{0.151} & \textbf{0.280} & \underline{0.178} & \underline{0.287} & \textbf{0.130} & \textbf{0.194} & 0.323 & 0.583 & 0.204 & 0.538 \\
\midrule


Model Knowledge
& \underline{0.160} & 0.303 & 0.190 & 0.311 & 0.157 & 0.237 & 0.330 & 0.579 & \textbf{0.191} & \textbf{0.443} \\
Movie Reviews
& 0.169 & 0.326 & 0.197 & 0.318 & 0.176 & 0.263 & 0.315 & 0.592 & \underline{0.203} & \underline{0.506} \\
Shopping History
& 0.163 & 0.312 & 0.189 & 0.301 & 0.160 & 0.240 & \textbf{0.309} & \underline{0.532} & 0.208 & 0.530 \\
SocSci210 MP3DR
& 0.186 & 0.352 & 0.203 & 0.324 & 0.168 & 0.248 & -- & -- & 0.251 & 0.746 \\
SocSci210 NJ5DX
& 0.192 & 0.366 & 0.204 & 0.331 & 0.172 & 0.247 & 0.364 & 0.781 & -- & -- \\

GSS
& -- & -- & 0.204 & 0.332 & 0.149 & 0.222 & 0.335 & 0.614 & 0.213 & 0.574 \\

OpinionQA W26
& 0.167 & 0.320 & -- & -- & 0.151 & 0.229 & 0.320 & 0.608 & 0.205 & 0.533 \\

OpinionQA W29
& 0.184 & 0.346 & 0.195 & 0.315 & -- & -- & 0.351 & 0.760 & 0.209 & 0.548 \\
\bottomrule

\end{tabular}
\caption{Survey simulation results on the target survey datasets in the low data-access regime. The best-performing simulation approach for each target survey is indicated in \textbf{bold}, with the second-best \underline{underlined}.}
\label{tab:out_domain_claude37}
\end{table*}

%% file: 5.experiments.tex
\section{Results}

We evaluate persona-conditioned simulation across five target survey datasets.~\footnote{The simulation is executed with Claude 3.7 Sonnet using $N_A=30$ agents for each demographic group. We provide additional results with open-weight LLMs in Appendix~\ref{app:additional_exp}.}
For each persona source dataset, we first construct a persona pool, then condition simulation on individual personas, and evaluate performance at the demographic group level.~\footnote{Note that the same persona pool can be reused for different target datasets in the out-of-domain setting.}

\subsection{Low-access Regime / Out-of-domain Setting}
\label{sec5.1_ood}
We begin with a practically motivated low-access setting in which only human behavior traces from non-target sources are available (RQ1). 
We consider persona sources including external behavioral corpora (movie reviews and shopping histories) and survey respondent data from other datasets/waves.

As shown in Table~\ref{tab:out_domain_claude37}, simulation performance varies across target survey datasets, partly reflecting differences in the average number of response options (SocSci210 subsets have an average of 7 response options and correspondingly result in highest TVD, whereas GSS has an average of 3.1 options and yields the lowest TVD). 
In addition, variation in survey topics may further contribute to performance differences.~\footnote{TVD and WD generally produce similar method rankings; unless otherwise noted, we therefore base our analysis on TVD.}


Personas induced from diverse out-of-domain sources generally do not outperform simple baselines that provide minimal context about target demographic groups. 
In particular, B3, which conditions each agent on a single attribute at inference time, remains consistently competitive.
Notably, personas derived from sources such as shopping histories and movie reviews are also not particularly effective.
One possible explanation is that persona sources often focus on relatively narrow topics (e.g., OpinionQA W29 centers on gender-related issues, while movie reviews reflect domain-specific preferences), leading to personas that are overly specific and misaligned with the target surveys.
This suggests that, without strong alignment between source behavioral data and the target population, additional persona detail may introduce noise rather than improve simulation alignment. Overall, the results indicate that group-level response distributions can often be approximated from minimal demographic context, and that naively incorporating out-of-domain personas does not reliably add predictive signal.

These results should also be interpreted in light of what distribution alignment does not capture. 
Even when persona-conditioned simulation does not substantially improve simulation alignment, it can still offer practical advantages in interpretability and explainability. 
Because each simulated response is tied to a particular persona induced from behavioral data, the resulting predictions are easier to inspect and explain than those from single-attribute baseline. 
Moreover, using a pool of personas enables the simulator to capture variation within a demographic group, rather than representing that group with a single-attribute demographic information.

We attribute the limited gains from out-of-domain personas to two primary factors. 
First, behavioral data from a source dataset may contain signals that are not useful for simulating users' opinions in target dataset. 
Second, our pipeline maps personas to target groups using demographic attributes inferred from behavioral data; inaccurate demographic attribute inference can therefore cause persona--group misassignment (e.g., assigning a persona derived from behavioral data of a female user to a male target demographic group because of incorrect demographic feature extraction), degrading simulation alignment.~\footnote{Details of demographic feature inference accuracy are provided in Tables~\ref{tab:conf_acc} and ~\ref{tab:demo_acc} in Appendix~\ref{app:extract_acc}.}
These observations motivate RQ2.
\input{tables/table_ood2}

\subsection{Mid-access Regime / Out-of-domain Setting with GT Demographics}
\label{sec5.2_ood_gt}
For datasets where individual-level GT demographic attributes are available, we evaluate two strategies for leveraging GT to improve persona-conditioned survey simulation: (i) \emph{GT assignment}, where GT  demographic information is used only to improve the assignment of personas to target demographic groups, and (ii) \emph{GT generation}, where persona generation is conditioned on both behavioral data and the corresponding GT demographic information about the user.

As shown in Table~\ref{tab:out_domain_claude37_2}, access to GT demographics substantially improves persona-conditioned simulation. 
GT assignment consistently reduces TVD relative to using the same persona source without GT demographic information, indicating that demographic attribute misassignment is a major contributor to reduced cross-domain simulation alignment.
In contrast, GT generation yields mixed effects: it can further reduce TVD in some settings, but it is not uniformly better than GT assignment, suggesting that conditioning persona generation on demographic information can introduce additional variance depending on the source behavioral data. 

Our qualitative analysis further suggests that conditioning persona generation on GT demographics can make the resulting personas disproportionately focused on surface-level attributes, rather than higher-level traits such as interests, preferences, or behavioral tendencies
(see Section~\ref{sec6_gt_surface}). 
Overall, these findings highlight accurate demographic assignment as a key bottleneck: when personas are correctly assigned to target demographic groups, persona-conditioned simulation becomes substantially more aligned to human responses even when the source behavioral data differs from the target survey population.

\subsection{High-access Regime / In-domain Setting}
\label{sec5.3_id}
We investigate how well personas induced from target-survey data generalize within the same domain (RQ3).
We use the GSS dataset for this setting, as it covers diverse topics and includes enough questions to allow meaningful splits.
Questions are divided into two subsets, with \{20\%, 40\%, 60\%, 80\%\} used for persona generation and the remainder for simulation evaluation.
\input{figures/figure3}

As shown in Figure~\ref{figure3}, simulation performance improves as more question history is used, with the largest gains at 80\%. 
One interpretation is that additional questions provide richer behavioral evidence, enabling the persona generator to infer more stable attributes, such as preferences or attitudes toward topics related to the held-out questions, which in turn improves response simulation.
Moreover, when personas incorporate a larger portion of each respondent's history, the simulator can better capture consistent response patterns across related questions. This leads to more faithful simulations of the respondent represented by each persona and, consequently, better alignment between human and agent response distributions within specific demographic groups.

In-domain personas with GT generation consistently outperforms the baseline once 40\% or more of the question history is used. 
This result suggests that personas generalize effectively across different question sets within the same domain, and further indicates that access to a sufficient amount of behavioral data from the target source is important for improving survey simulation.

To verify that this finding is not an artifact of a single data split, we 
repeat the experiment across multiple splits. GT Generation outperforms B3 in 
all three independent runs at the 80\% split (0.111 vs.\ 0.131; 0.139 
vs.\ 0.147; 0.153 vs.\ 0.169). 
While marginal standard deviations overlap 
between conditions (see Figure~\ref{figure3_split} for full results 
including standard deviations), this paired comparison confirms that the 
observed ranking is stable across splits. 
Due to computational cost, remaining 
conditions in Tables~\ref{tab:out_domain_claude37},~\ref{tab:out_domain_claude37_2}, and~\ref{tab:out_domain} are run 
once; we treat this repeated-run result as verification that our pipeline 
produces reproducible relative orderings.

%% file: tables/table_ood2.tex
\begin{table*}[t]
\centering
\scriptsize
\renewcommand{\arraystretch}{1.1}
\small
\setlength{\tabcolsep}{4pt}
\begin{tabular}{lcccccccccc}
\toprule

\multirow{2}{*}{Baseline/Persona Source}
& \multicolumn{2}{c}{GSS}
& \multicolumn{2}{c}{OpinionQA W26}
& \multicolumn{2}{c}{OpinionQA W29}
& \multicolumn{2}{c}{SocSci210 (MP3DR)}
& \multicolumn{2}{c}{SocSci210 (NJ5DX)} \\
\cmidrule(lr){2-3} \cmidrule(lr){4-5} \cmidrule(lr){6-7} \cmidrule(lr){8-9} \cmidrule(lr){10-11}
& TVD & WD & TVD & WD & TVD & WD & TVD & WD & TVD & WD \\
\midrule

B3 & \textbf{0.151} & \textbf{0.280} & 0.178 & 0.287 & \textbf{0.130} & \textbf{0.194} & 0.323 & 0.583 & 0.204 & 0.538 \\
\midrule


GSS
& -- & --
& 0.204 & 0.332
& 0.149 & 0.222
& 0.335 & 0.614
& 0.213 & 0.574 \\
\quad +GT assignment
& -- & --
& 0.200 & 0.325
& \underline{0.141} & \underline{0.210}
& 0.319 & \underline{0.512}
& 0.194 & 0.474 \\
\quad +GT generation
& -- & --
& 0.208 & 0.336
& 0.149 & 0.236
& \underline{0.317} & \textbf{0.482}
& 0.200 & 0.495 \\

OpinionQA W26
& 0.167 & 0.320
& -- & --
& 0.151 & 0.229
& 0.320 & 0.608
& 0.205 & 0.533 \\
\quad +GT assignment
& \underline{0.154} & 0.304
& -- & --
& \underline{0.141} & 0.214
& \textbf{0.305} & 0.544
& 0.192 & 0.472 \\
\quad +GT generation
& 0.157 & \underline{0.291}
& -- & --
& 0.150 & 0.234
& 0.329 & 0.571
& \textbf{0.186} & 0.457 \\

OpinionQA W29
& 0.184 & 0.346
& 0.195 & 0.315
& -- & --
& 0.351 & 0.760
& 0.209 & 0.548 \\
\quad +GT assignment
& 0.159 & 0.304
& \underline{0.172} & \underline{0.274}
& -- & --
& 0.326 & 0.648
& 0.188 & \textbf{0.447} \\
\quad +GT generation
& 0.159 & 0.293
& \textbf{0.169} & \textbf{0.267}
& -- & --
& 0.348 & 0.677
& \underline{0.187} & \underline{0.450} \\
\bottomrule

\end{tabular}
\caption{Survey simulation results on the target survey datasets in mid data-access regime. The best-performing approach for each survey is indicated in \textbf{bold}, with the second-best \underline{underlined}. \emph{GT assignment} uses GT demographics only for persona–group assignment, while \emph{GT generation} uses GT demographics during persona generation.}
\label{tab:out_domain_claude37_2}
\end{table*}

%% file: figures/figure3.tex
\begin{figure}[t]
\centering
\includegraphics[width= 0.95\columnwidth]{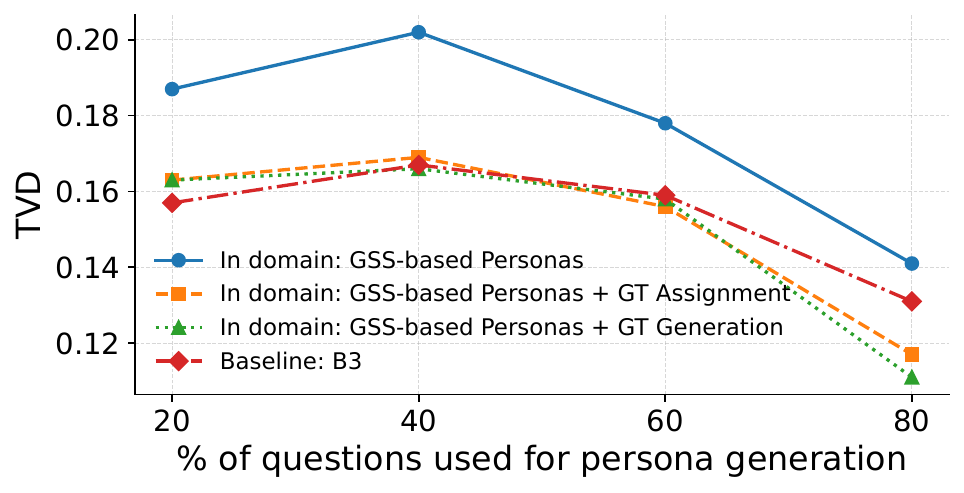} 
\caption{Survey simulation results in the in-domain setting, showing TVD for simulations using personas generated from varying fractions of GSS questions.}
\label{figure3}
\end{figure}

%% file: 6.discussion.tex
\input{figures/figure_demo}

\section{Discussion}
\label{sec6_ablation}
In this section, we provide deeper analyses of our persona-conditioned simulation approaches and baselines.
Unless stated otherwise, all analyses adopt the GSS evaluation configuration in which personas are induced from 80\% of the survey questions and evaluated on the remaining 20\% (as described in Section~\ref{sec5.3_id}).
We focus on representative methods from each investigated data-access regime spanning the single-attribute context baseline (B3), model-knowledge personas (Model Knowledge), out-of-domain personas with GT assignment sourced from OpinionQA Wave 26 (OpinionQA), and in-domain personas induced from 80\% of the question history with GT demographic information (GSS 80\%). 

\subsection{Effect of Persona Representativeness on Simulation Alignment}

We investigate whether simulation performance depends on how well the selected personas capture the ground-truth respondent population beyond the target attribute. 
Under the GSS 80\% setting, we compare, for each target subgroup (e.g., \texttt{Gender=male} or \texttt{Age=18--29}), the 30 personas used for simulation with the corresponding respondent subgroup.
For each non-target demographic attribute, we compute TVD between the attribute distribution across the selected personas and the distribution in the corresponding respondent group, considering all possible attribute values (e.g., education levels or income brackets). 
This yields a separate TVD score for each non-target attribute, which we then average to obtain an overall \emph{demographic mismatch} score for the demographic group.

As shown in Figure~\ref{figure_demo}, we observe a moderate positive association between demographic mismatch and simulation misalignment (TVD) (Pearson $=0.631$, Spearman $=0.323$), suggesting that demographic groups for which selected personas more closely resemble the corresponding respondent population tend to exhibit lower simulation misalignment. 
The weaker Spearman correlation, however, indicates that the relationship is not uniformly monotonic across demographic groups.

\input{figures/figure4}
\input{figures/figure5}

\subsection{Simulation Alignment Variations across Demographic Groups}
\label{sec6.1_demo_ablation}
\paragraph{Group size.}
We first compare performance across demographic-group sizes to assess whether simulation remains effective for relatively underrepresented groups across different data-access regimes.\footnote{The number of respondents in GSS does not fully reflect the true size of each demographic group, but it can serve as a proxy for population representation (e.g., smaller groups may be viewed as relatively underrepresented groups).}

As shown in Figure~\ref{figure4}, the ranking of methods is stable across all group sizes: personas induced from 80\% of GSS questions consistently achieve the best simulation alignment, followed by B3 and Model Knowledge, while OpinionQA exhibits the highest TVD.
This suggests that the benefits of in-domain persona induction are not limited to well-populated groups but extend to both sparse and large groups.
The advantage is especially pronounced for smaller groups, where underrepresentation presents a greater challenge.
Overall, these results suggest that in-domain persona induction yields stronger and more stable alignment across groups of varying sizes.




\paragraph{Demographic key.}
We next compare performance across demographic keys (e.g., \texttt{AGE} and \texttt{REGION}) to examine which attributes are more consistently captured by each method.
As shown in Figure~\ref{figure5}, GSS 80\% performs best across all demographic attributes, with uniformly low TVD values.
This indicates that personas induced from in-domain evidence generalize reliably across attributes, rather than improving only a narrow subset of groups.
Among the remaining methods, B3 is generally the strongest baseline, particularly on age, education, race, and sex, where differences are relatively small.
By contrast, OpinionQA consistently underperforms, indicating limited cross-domain simulation alignment when the source distribution diverges from the target.
Model Knowledge performs between B3 and OpinionQA, and overall shows notable degradation on POLPARTY (0.154), suggesting that political affiliation is especially difficult to simulate from model knowledge alone when the target distribution differs from the model's internal knowledge.

%% file: figures/figure_demo.tex
\begin{figure}[t]
\centering
\includegraphics[width= 0.95\columnwidth]{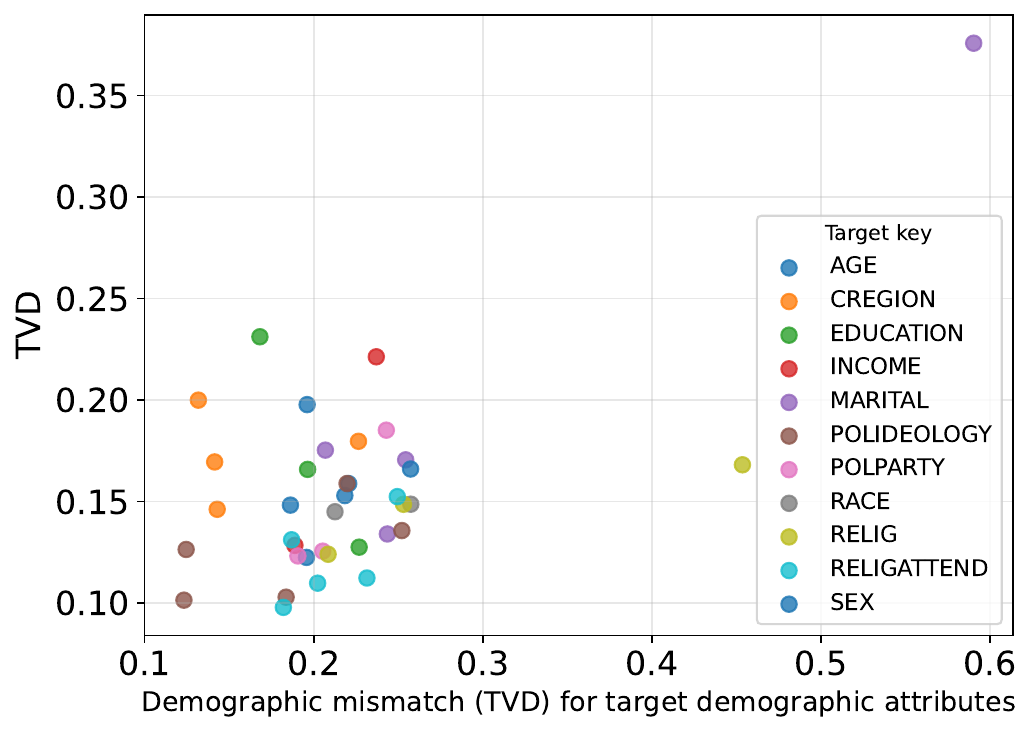} 
\caption{For each target demographic group, we measure the correlation between demographic mismatch score (as measured by TVD between selected personas and actual demographics) and survey TVD. Correlation coefficient: (Pearson=0.631, Spearman = 0.323)}
\label{figure_demo}
\vspace{-4mm}
\end{figure}

%% file: figures/figure4.tex
\begin{figure}[t]
\centering
\includegraphics[width= 0.95\columnwidth]{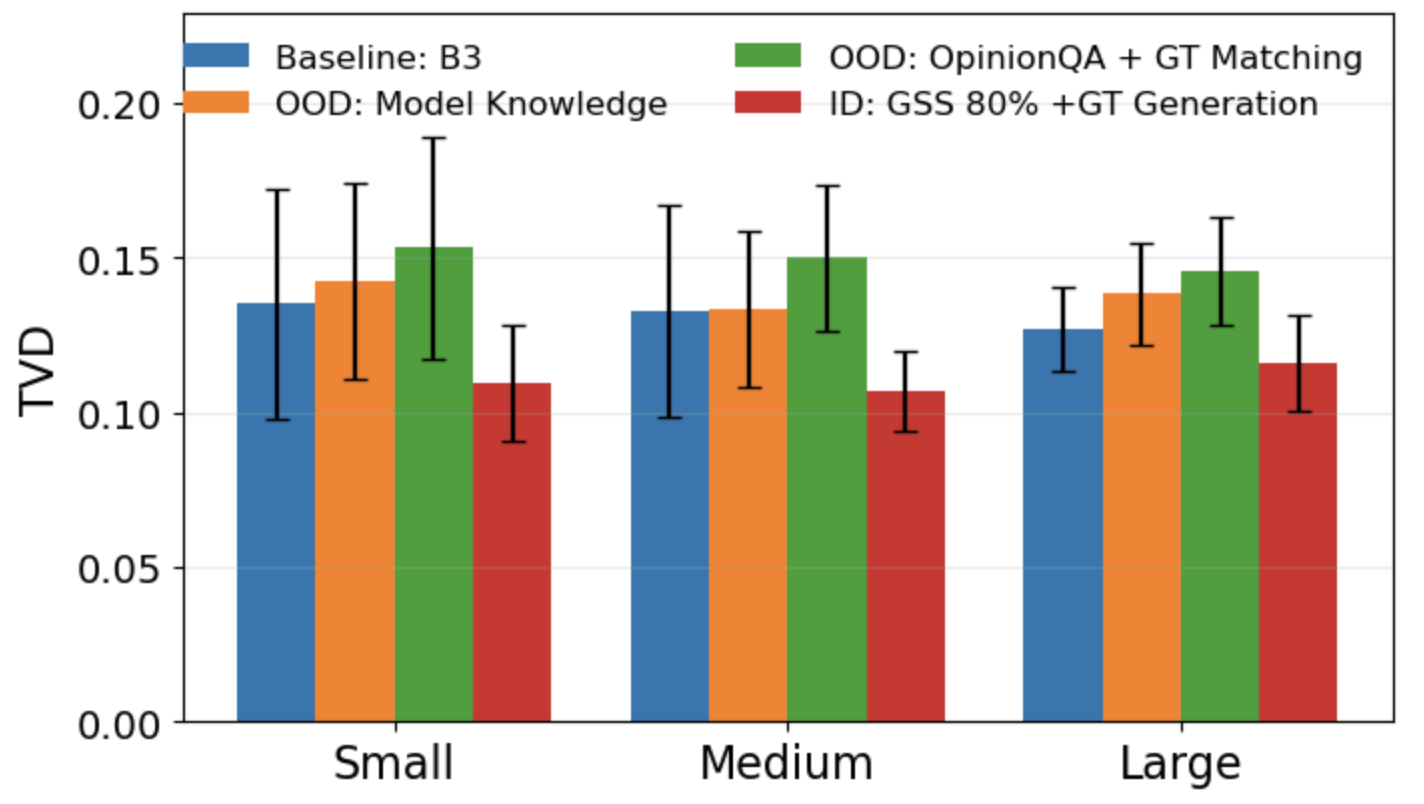} 
\caption{TVD across different demographic-group sizes after binning groups into Small, Medium, and Large buckets by respondent count in GSS dataset.}
\label{figure4}
\end{figure}

%% file: figures/figure5.tex
\begin{figure*}[t]
\centering
\includegraphics[width= 2\columnwidth]{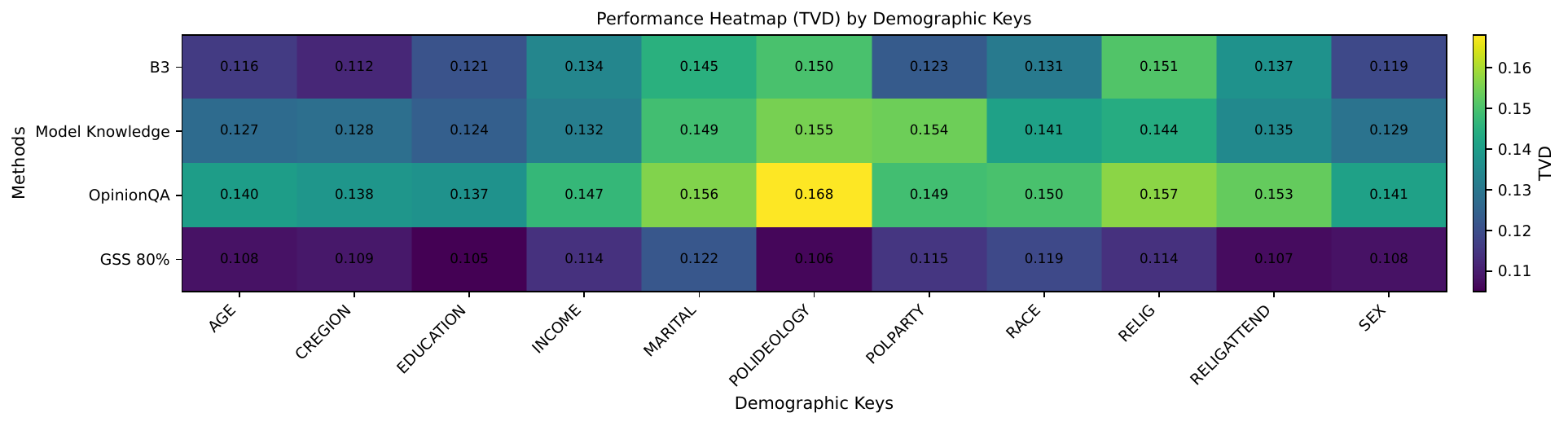} 
\caption{Performance Heatmap (TVD) across demographic keys}
\label{figure5}
\end{figure*}

%% file: 7.conclusion.tex
\section{Conclusion}
We presented a systematic study of persona-conditioned survey simulation under different data-access regimes, evaluated at the demographic-group level. 
Our results show that out-of-domain personas rarely outperform a single-attribute context baseline, primarily due to domain mismatch and errors in demographic attribute assignment. 
Their effectiveness improves substantially when ground-truth demographics are available, particularly for persona–group assignment.
In-domain personas perform progressively better as more question history is incorporated, suggesting that richer behavioral evidence enables more stable trait inference and better generalization to held-out questions. 
Overall, these findings highlight a practical insight for survey simulation: the critical factor is not merely the use of personas, but ensuring they are grounded in sufficient behavioral evidence and accurately matched to the target population.

%% file: limitation.tex
\section*{Limitations}
First, our evaluation has several scope limitations: it focuses on single-attribute demographic groups rather than intersectional groups, which may understate the difficulty of simulating finer-grained subpopulations; it is limited to U.S. respondents because our behavioral data sources are U.S.-centric; and it evaluates distribution-level alignment rather than individual-level, prioritizing privacy-aligned deployment. 
Extending the evaluation to intersectional groups and other regions, and connecting distributional accuracy to downstream decision-making and application-specific utility, remain important directions for future work.

Second, we did not investigate methods for improving demographic feature extraction accuracy. 
Table~\ref{tab:demo_acc} shows that extraction errors are a major bottleneck in the low-access regime, as they lead to persona misassignment and degrade simulation alignment. 
Our mid-access results suggest that accurate demographic assignment is the key factor, providing a concrete target for future work on more robust demographics inference methods.

Finally, due to computational cost, most experiments are run only once. Rather than focusing on stability through repeated runs, we prioritize generalizability by evaluating across 5 target datasets, 8 persona sources, and 3 LLMs, where consistent trends are observed. 
Since this is the first work to characterize persona generation from diverse behavioral data domains and systematically measure persona fidelity across domains, we prioritize generalization over stability.
For the in-domain setting, where uncertainty estimation is most relevant given the varying question history splits, we conduct multiple runs and report standard deviation in Figure~\ref{figure3_split} and Table~\ref{tab:iterations}.

%% file: ethic.tex
\section*{Ethical Considerations}
In our experiments, we utilize publicly available datasets, including OpinionQA~\cite{santurkar2023whose}, GSS~\cite{davern2024gss}, SocSci210~\cite{kolluri2025finetuning}, movie review~\cite{yoon2024evaluating}, and shopping histories~\cite{berke2024open}.
These datasets are widely adopted and well-established within the research community, and their use raises no additional privacy or consent concerns beyond those addressed in the original dataset releases.

All large language models used in this work were accessed through their official and publicly available sources.
Our use of these models complies with their stated terms of service and aligns with open science and reproducibility principles.

During the preparation of this manuscript, we employed an AI-assisted writing tool at the sentence level to support drafting and linguistic refinement.

Beyond the privacy safeguards already described, we note additional societal risks specific to persona-conditioned survey simulation. 
First, tools that simulate group-level opinion distributions could be misused for political microtargeting or opinion manipulation. 
Second, survey simulation risks being used as a substitute for genuine stakeholder engagement, particularly for underrepresented or hard-to-reach populations. 
Our results suggest this is especially risky in the low-access regime, where simulated distributions can appear plausible while diverging substantially from real subgroup opinions due to demographic misassignment (Section~\ref{sec5.1_ood}). 
We emphasize that our findings argue against using out-of-domain persona simulation as a drop-in replacement for genuine data collection, and recommend that any deployment report demographic-assignment accuracy and group-level sample size alongside simulated distributions.

%% file: appendix.tex
\input{tables/table_dataset}
\input{figures/figure3_split}

\input{tables/table_gt_vs_free}
\input{tables/table_out}
\input{tables/table_iterations}
\section{Pilot Study}
\label{app:pilot}

In our pilot study, we explore several design choices for using synthetic personas in survey-response simulation.

\subsection{Verbalized Distribution vs. Option Selection}
\label{sec_option_selection}

Section~\ref{sec_3.1} elicits a \emph{verbalized} probability distribution from each persona-conditioned agent (e.g., \texttt{A: 30\%, B: 70\%}) and then averages these distributions. 
As an alternative, we can estimate the group-level distribution by prompting each agent to select \emph{a single} answer option from the available set and then aggregating these selections across agents.

Concretely, for each agent $j \in \{1,\ldots,N_A\}$ conditioned on persona $P_j$, we prompt the LLM to output exactly one option $\hat{x}^{(j)}_{q} \in \mathcal{X}_q$ for question $q$. 
We then form an empirical distribution by counting how often each option is selected:
\begin{equation}
S_{q,g}(x)
=\frac{1}{N_A}\sum_{j=1}^{N_A}\mathbbm{1}\!\left\{\hat{x}^{(j)}_{q}=x\right\},
\quad x \in \mathcal{X}_q,
\end{equation}
where $\mathbbm{1}\{\cdot\}$ denotes the indicator function.

\subsection{Number of Personas and Aggregation Scheme: Verbalized Distributions vs.\ Option Selection}

We compare two aggregation schemes: (i) \emph{verbalized distribution} outputs, in which the simulator directly produces a probability distribution over answer options, and (ii) \emph{option selection + aggregation}, in which each persona selects a single option and responses are aggregated by voting. 
We also vary the number of personas used for each demographic group.

We conduct experiments on SimBench-OpinionQA~\cite{hu2025simbench} using Claude 3.7 Sonnet and Llama-3.1-70B as simulators. 
For each target demographic value, we evaluate simulations using 30, 50, and 100 personas, and compare the two aggregation schemes.

As shown in Figure~\ref{fig:num_persona}, verbalized distributions consistently outperform option selection + aggregation. 
This finding aligns with prior work suggesting that direct distribution elicitation better preserves minority probability mass than hard voting~\cite{meister2025benchmarking}. 
We also find that performance largely saturates beyond 30 personas: once the number of personas exceeds approximately 30, the improvement curve plateaus, suggesting diminishing returns from adding more agents under this setup.

\input{figures/figure6}

\subsection{Why Does Option Selection Underperform?}

We next investigate why the option selection scheme performs worse than verbalized distributions. 
Our working hypothesis is that majority voting over single-choice responses produces overly peaked distributions that over-represent the dominant preference while systematically under-representing minority opinions.

To examine this hypothesis, we compute two diagnostics: 
(1) the average Shannon entropy of the simulated response distribution, where lower entropy indicates a more concentrated distribution, and 
(2) the number of non-chosen options across questions, where for each of the 964 survey questions we count how many answer options receive zero selections under option selection + aggregation.

\input{figures/figure7}
\input{figures/figure8}

Across questions, we find that option selection frequently collapses to a single dominant option, with most of the probability mass concentrated on one answer and many alternatives never selected even once. 
As a result, the simulated distribution primarily reflects the majority view and fails to capture the within-group heterogeneity observed in human data.

This effect is further amplified by the persona construction used in this pilot setting: personas are generated from only a single demographic attribute (e.g., \textit{Gender: female}). 
With such limited information to differentiate personas, agents tend to converge on the same ``typical'' response, and voting-based aggregation then amplifies this consensus while suppressing minority mass. 
This mechanism plausibly explains why option-selection-based distributions align less well with the true subgroup-level response distributions.

\subsection{Persona Induction Strategies}
\label{app:persona_prompt}
To study how prompting affects persona quality, we compare four induction strategies:
\begin{enumerate}
    \item \textbf{Free.} Generate a free-form bio from the individual's behavioral data only.
    \item \textbf{Free+GT.} Generate a free-form bio from the individual's behavioral data + Ground-truth demographic feature as described in Section~\ref{sec5.2_ood_gt}.
    \item \textbf{Target-aware.} Provide brief context about the target survey datasets to be simulated (task description, domains, and example questions), then generate a bio.
    \item \textbf{Rubric-based.} Ask the generator to answer survey-style rubric questions (aligned with target survey domains) based on the individual's data, then synthesize a bio.
\end{enumerate}
\input{figures/app_prompt_free}
\input{figures/app_prompt_freegt}
Each persona includes (i) a short bio and (ii) extracted demographic features with confidence scores. 
Personas are then grouped by demographic features (e.g., Gender: male) and selected accordingly for subgroup simulation.
In pilot experiments, we found that using the \textit{Free} prompt template for persona generation and selecting personas with high confidence on the target demographic attribute consistently yielded the best performance.
Detailed prompt templates used in the experiments (Free, Free+GT) are described in Figure~\ref{fig:app_free_persona} and ~\ref{fig:app_freegt_persona}, respectively.




\section{Persona Detail}
\label{app:persona}
Each key has a closed set of possible values:
\begin{itemize}
    \item \textbf{Region}: closed options (South \textbar\ Northeast \textbar\ West \textbar\ Midwest)
    \item \textbf{Age}: closed options (18--29 \textbar\ 30--49 \textbar\ 50--64 \textbar\ 65+)
    \item \textbf{Sex}: closed options (male \textbar\ female)
    \item \textbf{Education}: closed options (high school graduate \textbar\ college graduate/some postgrad \textbar\ some college, no degree \textbar\ associate's degree \textbar\ postgraduate \textbar\ less than high school)
    \item \textbf{Marital}: closed options (married \textbar\ divorced \textbar\ widowed \textbar\ living with a partner \textbar\ never been married)
    \item \textbf{Religion}: closed options (protestant \textbar\ roman catholic \textbar\ atheist \textbar\ agnostic \textbar\ nothing in particular)
    \item \textbf{Religious Attendance}: closed options (more than once a week \textbar\ seldom \textbar\ never \textbar\ a few times a year \textbar\ once or twice a month \textbar\ once a week)
    \item \textbf{Political Party}: closed options (Democrat \textbar\ Independent \textbar\ Other \textbar\ Republican)
    \item \textbf{Income}: closed options (Less than \$30{,}000 \textbar\ \$30{,}000--\$50{,}000 \textbar\ \$50{,}000--\$75{,}000 \textbar\ \$75{,}000--\$100{,}000 \textbar\ \$100{,}000 or more)
    \item \textbf{Political Ideology}: closed options (liberal \textbar\ moderate \textbar\ conservative \textbar\ very conservative \textbar\ very liberal)
    \item \textbf{Race}: closed options (black \textbar\ hispanic \textbar\ asian \textbar\ white)
\end{itemize}

\section{Metric Details}
\label{app:metric}
TVD measures the $\ell_1$ discrepancy between two discrete distributions:

\vspace{-4mm}
\begin{equation}
\small
\mathrm{TVD}\!\left(U_{i,g}, S_{i,g}\right)
=\frac{1}{2}\sum_{k=1}^{N_{\mathrm{option},i}} \left| U_{i,g}(x_k)-S_{i,g}(x_k)\right|.
\end{equation}
\noindent
WD captures the minimal ``mass transport'' needed to transform one distribution into another while respecting the ordering of options.\footnote{This metric complements TVD by accounting for the ordinal structure of response options. All target datasets used in our experiments consist of survey questions with ordinal options, which are suited for WD.}
\begin{equation}
\small
W_1\!\left(U_{i,g}, S_{i,g}\right)
=
\sum_{k=1}^{N_{\mathrm{option},i}}
\left|
\sum_{r=1}^{k} U_{i,g}(x_r)
-
\sum_{r=1}^{k} S_{i,g}(x_r)
\right|.
\end{equation}

\section{Dataset Details}
\label{app:dataset}
\subsection{Statistics of Target Datasets and Persona-source Datasets}
As discussed in Section~\ref{sec_4}, we utilize 5 different survey datasets for test and use various sources of persona induction. Table~\ref{tab:dataset_stat} summarize the statistics for datasets used in our experiments.

\subsection{Group-level Test Instances Example}
\label{app:test_example}
\input{figures/app_instance_example}
Figure~\ref{fig:app_test_example} shows example test instances from each target survey datasets.

\section{Inference Details}
For persona generation we used temperature = 0.7 and for simulation inference (both baseline and persona conditioned inference) we used temperature = 0.0

Prompt used for synthetic persona-conditioned simulation is described in Figure~\ref{fig:app_inference_prompt}.
\input{figures/app_inference_prompt}

\section{Accuracy of Demographic Feature Extraction in Persona Generation}
\label{app:extract_acc}
In this section, we evaluate whether demographic features extracted from behavior data match ground-truth demographics.
Ground-truth demographics are available for OpinionQA W26/W29
and GSS.

\subsection{Extraction Accuracy by Confidence Level}
We first analyze how extraction accuracy varies with the model-assigned confidence level. 
As shown in Table~\ref{tab:conf_acc}, accuracy usually increases with confidence across datasets, consistent with Appendix~\ref{app:pilot}'s finding that high-confidence persona selection improves simulation performance.
\input{tables/tab_conf_acc}

We then test whether the personas used for simulation ($N_A=30$
per demographic feature) are actually matched to the intended target group. 
The results shown in Table~\ref{tab:demo_acc} indicate substantial mismatch, which helps explain the weaker performance of personas constructed without ground-truth demographic conditioning: when persona demographics are misaligned with the target subgroup, persona conditioning can become noisy or misleading.
\input{tables/table_demo_acc}

\section{Additional Results}
\label{app:additional_exp}
To test robustness of survey simulation across model families, we report results for two additional models (Llama-3.3-70B~\cite{dubey2024llama} and Qwen-3-32B~\cite{yang2025qwen3}) using personas induced from diverse sources in Table~\ref{tab:out_domain}. The findings in Section~\ref{sec5.1_ood} regarding persona utility hold consistently across models.

\section{Additional Discussions}
\label{app:discussion}
\subsection{Personas Induced from Behavioral Data Alone vs. with GT Demographics }
\label{sec6_gt_surface}
Providing GT demographics during persona generation sometimes reduces TVD in certain settings, but it does not uniformly outperform GT assignment, as discussed in Section~\ref{sec5.2_ood_gt}. 
Our manual inspection of these personas suggests a systematic difference in how they are constructed. 
Personas generated with GT demographics typically begin with an explicit listing of the provided demographic attributes, followed by a behavioral summary grounded in the respondent's observed answers. 
In contrast, personas generated without GT demographic information often express demographic characteristics only implicitly, or leave them underspecified, while relying more heavily on generic behavioral patterns (see Appendix Table~\ref{tab:gt_feature_free} for an example).

These observations suggest that providing GT demographic information may encourage the generation of shallower personas that focus less on actual behavioral patterns. 
This may help explain why GT-informed persona generation does not consistently lead to improved simulation performance.

\subsection{Broader Implications for Survey Simulation Systems}
Taken together, our results suggest that the main value of synthetic personas for survey simulation is not simply that they provide richer prompt text, but that they can encode behaviorally grounded evidence when they are representative of the population being simulated.
In practice, this means future survey simulation systems should treat persona use as an alignment problem rather than a prompting problem: they should validate how well personas represent source-population and explicitly audit persona–group matching.
Our findings also suggest that when GT demographics are available, they are especially useful for matching, whereas injecting them directly into persona generation may encourage shallower, surface-level profiles. 
Finally, when partial target-survey history is available, systems should prioritize it, since richer in-domain behavioral evidence yields more stable traits, better transfer to unseen questions, and more consistent gains across demographic groups. 
The practical impact of this work is therefore to provide design guidance for building more reliable and privacy-aligned survey simulators using synthetic personas derived from diverse behavioral data.

%% file: tables/table_dataset.tex

\begin{table}[t]
\centering
\scriptsize
\setlength{\tabcolsep}{3pt}
\renewcommand{\arraystretch}{1.05}

\begin{tabularx}{0.95\columnwidth}{@{}l >{\raggedright\arraybackslash}X r r r@{}}
\hline
\multicolumn{5}{@{}l}{\textbf{Persona Source}}\\
\hline
\textbf{Dataset} & \textbf{Domain} &  \textbf{\#Q}   & \textbf{Avg. Resp.}& \textbf{\#Resp.}\\
\hline
GSS    & General Survey           &  60  & 36.1&3309 \\
OpinionQA W26 & Survey: Guns                      & 78 & 46.2& 4168 \\
OpinionQA W29 & Survey: Gender                   & 77  & 53.8 & 4867\\
SocSci210 MP3DR  & Survey: Gender                    & 39 & 6.9 & 4161 \\
SocSci210 NJ5DX  & Survey: Inequality        & 48  & 18.2 & 1814\\
IMDB   & Movie review              & -- & 21.2& 1079 \\
E-commerce & Shopping history          & -- & 368.2 & 5027\\
\hline
\multicolumn{5}{@{}l}{\textbf{Test Dataset}}\\
\hline
\textbf{Dataset} & \textbf{\#Demog. feats.} & \textbf{\#Q} & \textbf{\#Opt.} & \textbf{Total}\\
\hline
GSS    & 40 & 60 & 3.1 & 2280 \\
OpinionQA W26 & 50 & 78 & 4.5 & 3474 \\
OpinionQA W29 & 50 & 77 & 4.1 & 3622 \\
SocSci210 MP3DR  & 18 & 39 & 7.0 & 555  \\
SocSci210 NJ5DX  & 25 & 44 & 7.0 & 934  \\
\hline
\end{tabularx}

\caption{Summary statistics of persona sources and test datasets.}
\label{tab:dataset_stat}
\end{table}

%% file: figures/figure3_split.tex
\begin{figure}[t]
\centering
\includegraphics[width= 0.95\columnwidth]{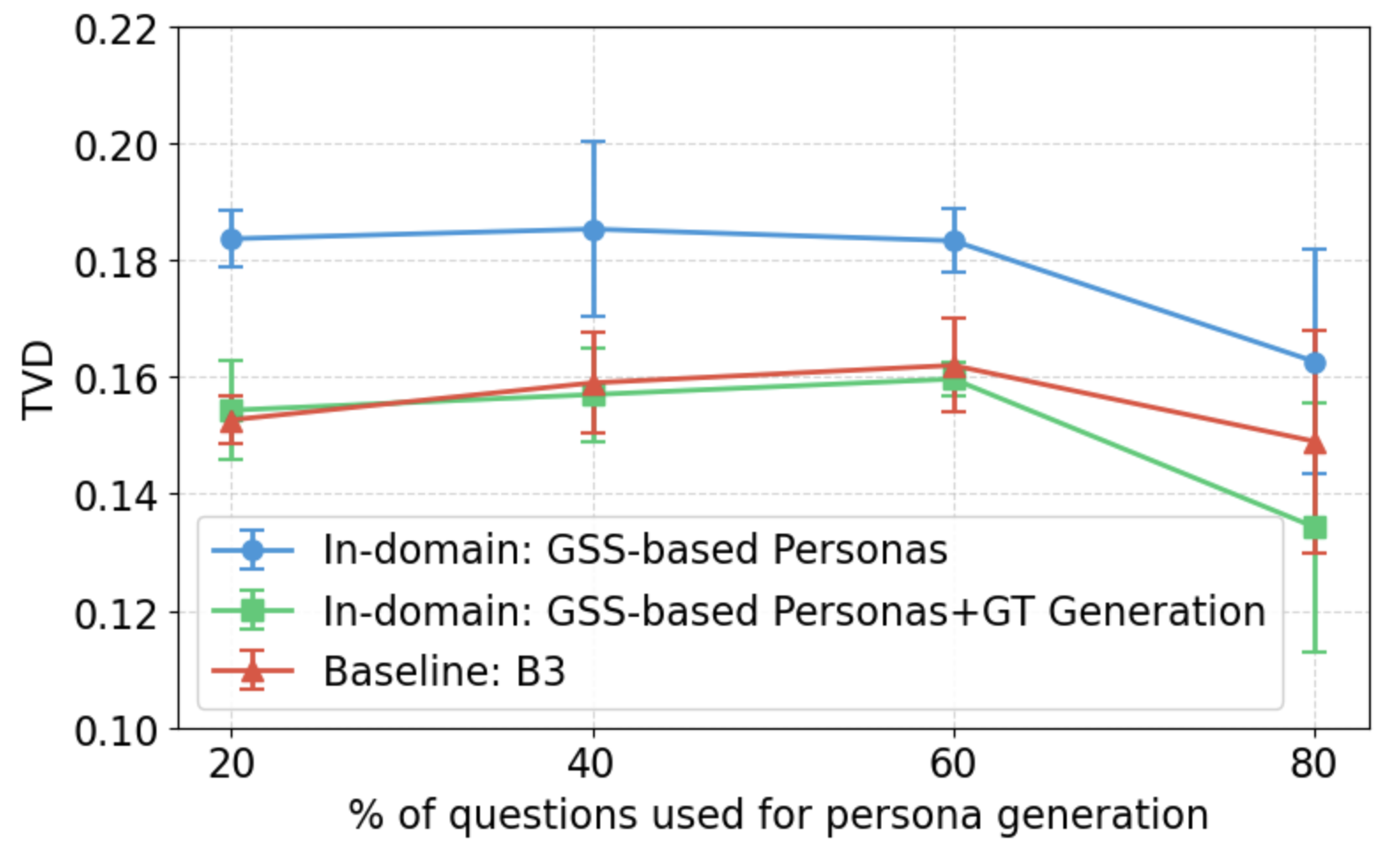} 
\caption{Survey simulation results in the in-domain setting, showing TVD and standard deviation (Std.) for simulations using personas generated from varying fractions of GSS questions. Results for each iteration and Std. are available in Table~\ref{tab:iterations}}
\label{figure3_split}
\vspace{-4mm}
\end{figure}

%% file: tables/table_gt_vs_free.tex

\begin{table*}[t]
\centering
\small
\setlength{\tabcolsep}{8pt}
\renewcommand{\arraystretch}{1.15}

\begin{tabularx}{\textwidth}{|>{\raggedright\arraybackslash}X|>{\raggedright\arraybackslash}X|}
\hline
\textbf{GT Generation} & \textbf{Free} \\
\hline
\hl{You are a 33-year-old Black man living in the Northeast who works in education administration at an elementary or secondary school. You're a college graduate with a bachelor's degree who has never married but has two children. You identify as Protestant and attend religious services weekly, indicating that faith is an important part of your life.}\par

Politically, you lean Democratic but consider yourself moderate rather than strongly liberal. You consistently vote in presidential elections, supporting Democratic candidates like Clinton and Biden. You have only some confidence in the executive branch of government, suggesting a degree of political skepticism despite your party affiliation...
&
You are a politically moderate Democrat who consistently votes in elections and supports Democratic presidential candidates. While you identify as middle-of-the-road politically, your policy positions tend to lean progressive on domestic issues - you support increased spending on education, healthcare, environmental protection, addressing urban problems, and improving conditions for Black Americans. You oppose the death penalty and believe premarital sex is not morally wrong.\par

You're fiscally more conservative on international matters, believing we spend too much on military and foreign aid. You're civically engaged and vote regularly in presidential elections...\\
\hline
\end{tabularx}

\caption{GT feature vs. Free description}
\label{tab:gt_feature_free}
\end{table*}

%% file: tables/table_out.tex
\begin{table*}[t]
\centering
\scriptsize
\renewcommand{\arraystretch}{1.1}
\small
\setlength{\tabcolsep}{4pt}
\begin{tabular}{l|cc|cc|cc|cc|cc}
\hline

\rowcolor{gray!10}
\multicolumn{11}{c}{\rule{0pt}{1.05em}\textbf{\textit{Simulator: Claude 3.7}}} \\
\hline

Target Data
& \multicolumn{2}{c}{GSS}
& \multicolumn{2}{c}{OpinionQA W26}
& \multicolumn{2}{c}{OpinionQA W29}
& \multicolumn{2}{c}{SocSci210 (MP3DR)}
& \multicolumn{2}{c}{SocSci210 (NJ5DX)} \\
\cline{2-11}
& TVD & WD & TVD & WD & TVD & WD & TVD & WD & TVD & WD \\
\hline

B1 & 0.169 & 0.322 & 0.189 & 0.299 & 0.171 & 0.262 & \underline{0.311} & 0.474 & 0.241 & 0.678 \\
B2 & \underline{0.160} & 0.301 & \textbf{0.173} & 0.274 & \underline{0.139} & 0.204 & 0.314 & 0.552 & 0.204 & 0.560 \\
B3 & \textbf{0.151} & 0.280 & \underline{0.178} & 0.287 & \textbf{0.130} & 0.194 & 0.323 & 0.583 & 0.204 & 0.538 \\
\hline

\rowcolor{gray!5}
\multicolumn{11}{l}{\textit{Persona Source}} \\
\hline

Model Knowledge
& \underline{0.160} & 0.303
& 0.190 & 0.311
& 0.157 & 0.237
& 0.330 & 0.579
& \textbf{0.191} & 0.443 \\
Movie
& 0.169 & 0.326
& 0.197 & 0.318
& 0.176 & 0.263
& 0.315 & 0.592
& \underline{0.203} & 0.506 \\
Shopping His
& 0.163 & 0.312
& 0.189 & 0.301
& 0.160 & 0.240
& \textbf{0.309} & 0.532
& 0.208 & 0.530 \\
Survey MP3DR
& 0.186 & 0.352
& 0.203 & 0.324
& 0.168 & 0.248
& -- & --
& 0.251 & 0.746 \\
Survey NJ5DX
& 0.192 & 0.366
& 0.204 & 0.331
& 0.172 & 0.247
& 0.364 & 0.781
& -- & -- \\
OpinionQA W26
& 0.167 & 0.320
& -- & --
& 0.151 & 0.229
& 0.320 & 0.608
& 0.205 & 0.533 \\
OpinionQA W29
& 0.184 & 0.346
& 0.195 & 0.315
& -- & --
& 0.351 & 0.760
& 0.209 & 0.548 \\
GSS
& -- & --
& 0.204 & 0.332
& 0.149 & 0.222
& 0.335 & 0.614
& 0.213 & 0.574 \\
\hline

\rowcolor{gray!10}
\multicolumn{11}{c}{\rule{0pt}{1.05em}\textbf{\textit{Simulator: Llama-3.3-70B}}} \\
\hline

Target Data
& \multicolumn{2}{c}{GSS}
& \multicolumn{2}{c}{OpinionQA W26}
& \multicolumn{2}{c}{OpinionQA W29}
& \multicolumn{2}{c}{SocSci210 (MP3DR)}
& \multicolumn{2}{c}{SocSci210 (NJ5DX)} \\
\cline{2-11}
& TVD & WD & TVD & WD & TVD & WD & TVD & WD & TVD & WD \\
\hline

B1 & 0.200 & 0.357 & 0.200 & 0.318 & 0.200 & 0.295 & 0.475 & 0.863 & 0.305 & 0.879 \\
B2 & \textbf{0.168} & 0.320 & \textbf{0.183} & 0.287 & \underline{0.188} & 0.279 & 0.456 & 0.861 & 0.325 & 1.030 \\
B3 & 0.171 & 0.311 & \textbf{0.183} & 0.291 & \textbf{0.181} & 0.264 & 0.431 & 0.808 & 0.311 & 0.947 \\
\hline

\rowcolor{gray!5}
\multicolumn{11}{l}{\textit{Persona Source}} \\
\hline

Model Knowledge
& \underline{0.169} & 0.320
& \underline{0.185} & 0.297
& 0.195 & 0.294
& 0.427 & 0.687
& \textbf{0.275} & 0.803 \\
Movie
& 0.191 & 0.362
& 0.210 & 0.341
& 0.239 & 0.366
& 0.472 & 0.868
& 0.315 & 0.983 \\
Shopping His
& 0.180 & 0.342
& 0.212 & 0.345
& 0.208 & 0.317
& 0.441 & 0.820
& \underline{0.277} & 0.864 \\
Survey MP3DR
& 0.209 & 0.392
& 0.199 & 0.322
& 0.235 & 0.348
& -- & --
& 0.390 & 1.290 \\
Survey NJ5DX
& 0.209 & 0.398
& 0.200 & 0.323
& 0.225 & 0.334
& \underline{0.398} & 0.787
& -- & -- \\
OpinionQA W26
& 0.192 & 0.355
& -- & --
& 0.208 & 0.309
& 0.453 & 0.832
& 0.283 & 0.860 \\
OpinionQA W29
& 0.194 & 0.366
& 0.204 & 0.327
& -- & --
& \textbf{0.384} & 0.660
& \textbf{0.275} & 0.835 \\
GSS
& -- & --
& 0.209 & 0.336
& 0.198 & 0.306
& 0.431 & 0.782
& 0.303 & 0.943 \\
\hline

\rowcolor{gray!10}
\multicolumn{11}{c}{\rule{0pt}{1.05em}\textbf{\textit{Simulator: Qwen3-32B}}} \\
\hline

Target Data
& \multicolumn{2}{c}{GSS}
& \multicolumn{2}{c}{OpinionQA W26}
& \multicolumn{2}{c}{OpinionQA W29}
& \multicolumn{2}{c}{SocSci210 (MP3DR)}
& \multicolumn{2}{c}{SocSci210 (NJ5DX)} \\
\cline{2-11}
& TVD & WD & TVD & WD & TVD & WD & TVD & WD & TVD & WD \\
\hline

B1 & 0.186 & 0.337 & 0.228 & 0.381 & 0.227 & 0.350 & 0.595 & 1.674 & 0.247 & 0.710 \\
B2 & 0.180 & 0.328 & 0.235 & 0.390 & \textbf{0.189} & 0.277 & 0.556 & 1.473 & 0.260 & 0.721 \\
B3 & 0.178 & 0.319 & 0.221 & 0.369 & 0.194 & 0.282 & 0.547 & 1.486 & 0.254 & 0.716 \\
\hline

\rowcolor{gray!5}
\multicolumn{11}{l}{\textit{Persona Source}} \\
\hline

Model Knowledge
& \textbf{0.173} & 0.311
& \textbf{0.207} & 0.346
& \underline{0.192} & 0.284
& 0.559 & 1.451
& \underline{0.219} & 0.561 \\
Movie
& 0.183 & 0.324
& 0.229 & 0.380
& 0.214 & 0.321
& 0.542 & 1.245
& 0.253 & 0.699 \\
Shopping His
& \underline{0.174} & 0.314
& 0.218 & 0.358
& 0.193 & 0.282
& \underline{0.523} & 1.306
& 0.226 & 0.626 \\
Survey MP3DR
& 0.199 & 0.366
& 0.228 & 0.376
& 0.212 & 0.310
& -- & --
& 0.284 & 0.846 \\
Survey NJ5DX
& 0.203 & 0.372
& 0.222 & 0.373
& 0.217 & 0.315
& 0.592 & 1.594
& -- & -- \\
OpinionQA W26
& 0.176 & 0.323
& -- & --
& \underline{0.192} & 0.283
& \textbf{0.492} & 1.200
& 0.229 & 0.621 \\
OpinionQA W29
& 0.184 & 0.341
& 0.217 & 0.363
& -- & --
& 0.609 & 1.738
& \textbf{0.218} & 0.586 \\
GSS
& -- & --
& \underline{0.215} & 0.361
& 0.198 & 0.291
& 0.535 & 1.254
& 0.229 & 0.643 \\
\hline

\end{tabular}
\caption{Survey simulation results using three different LLMs on the target survey datasets in the low data-access regime. The best-performing simulation approach for each target survey is indicated in \textbf{bold}, with the second-best \underline{underlined}.}
\label{tab:out_domain}
\end{table*}

%% file: tables/table_iterations.tex
\begin{table*}[h]
\centering
\begin{tabular}{clccc}
\toprule
\textbf{Iteration} & \textbf{Question Split} & \textbf{GSS Persona} & \textbf{+GT Generation} & \textbf{B3} \\
\midrule
\multirow{4}{*}{1}
 & 20\% & 0.187 & 0.163 & 0.157 \\
 & 40\% & 0.202 & 0.166 & 0.167 \\
 & 60\% & 0.178 & 0.158 & 0.156 \\
 & 80\% & 0.141 & 0.111 & 0.131 \\
\midrule
\multirow{4}{*}{2}
 & 20\% & 0.178 & 0.154 & 0.149 \\
 & 40\% & 0.173 & 0.151 & 0.150 \\
 & 60\% & 0.183 & 0.158 & 0.159 \\
 & 80\% & 0.169 & 0.139 & 0.147 \\
\midrule
\multirow{4}{*}{3}
 & 20\% & 0.186 & 0.146 & 0.152 \\
 & 40\% & 0.181 & 0.154 & 0.160 \\
 & 60\% & 0.189 & 0.163 & 0.171 \\
 & 80\% & 0.178 & 0.153 & 0.169 \\
\midrule
\multirow{4}{*}{\textbf{Mean $\pm$ STD}}
 & 20\% & $0.184 \pm 0.005$ & $0.154 \pm 0.009$ & $0.153 \pm 0.004$ \\
 & 40\% & $0.185 \pm 0.015$ & $0.157 \pm 0.008$ & $0.159 \pm 0.009$ \\
 & 60\% & $0.183 \pm 0.006$ & $0.160 \pm 0.003$ & $0.162 \pm 0.008$ \\
 & 80\% & $0.163 \pm 0.019$ & $0.134 \pm 0.021$ & $0.149 \pm 0.019$ \\
\bottomrule
\end{tabular}
\caption{Survey simulation results in the in-domain setting across iterations, showing TVD for simulations using personas generated from varying fractions of GSS questions.}
\label{tab:iterations}
\end{table*}

%% file: figures/figure6.tex
\begin{figure}[t]
\centering
\includegraphics[width= 0.95\columnwidth]{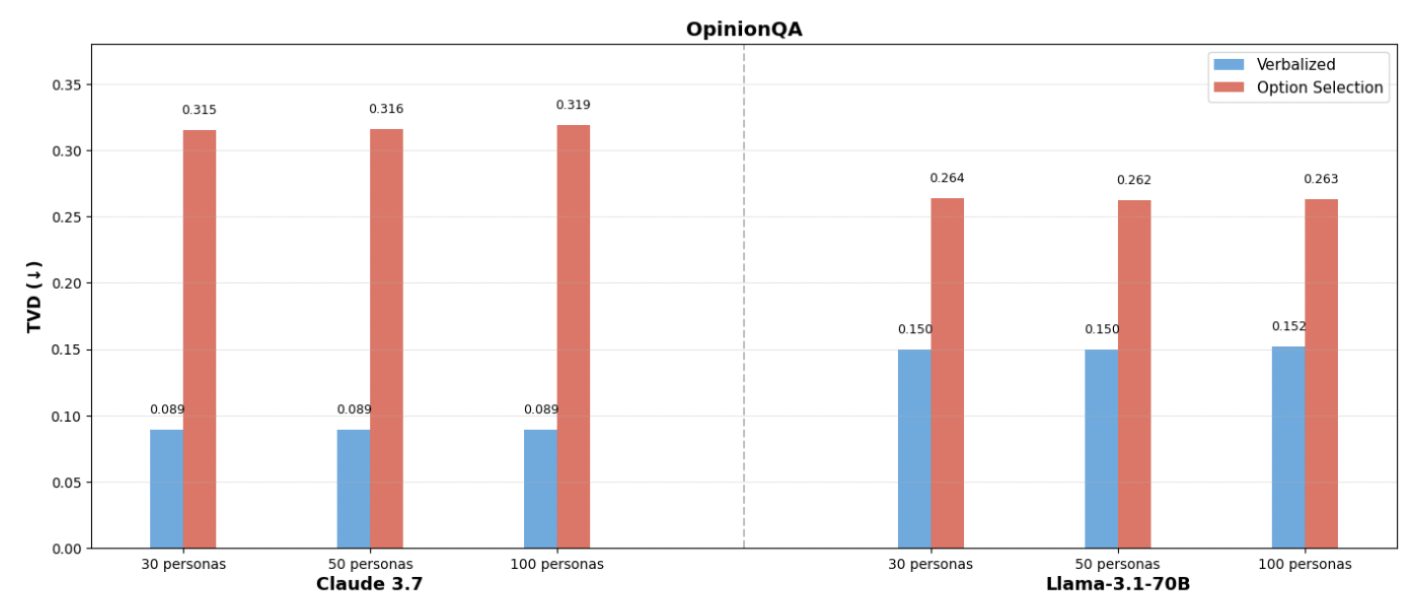} 
\caption{Performance of Claude-3.7-Sonnet and Llama-3.1-70B model on OpinionQA across different number of personas and different distribution aggregation methods}
\label{fig:num_persona}
\vspace{-4mm}
\end{figure}

%% file: figures/figure7.tex
\begin{figure}[t]
\centering
\includegraphics[width= 0.95\columnwidth]{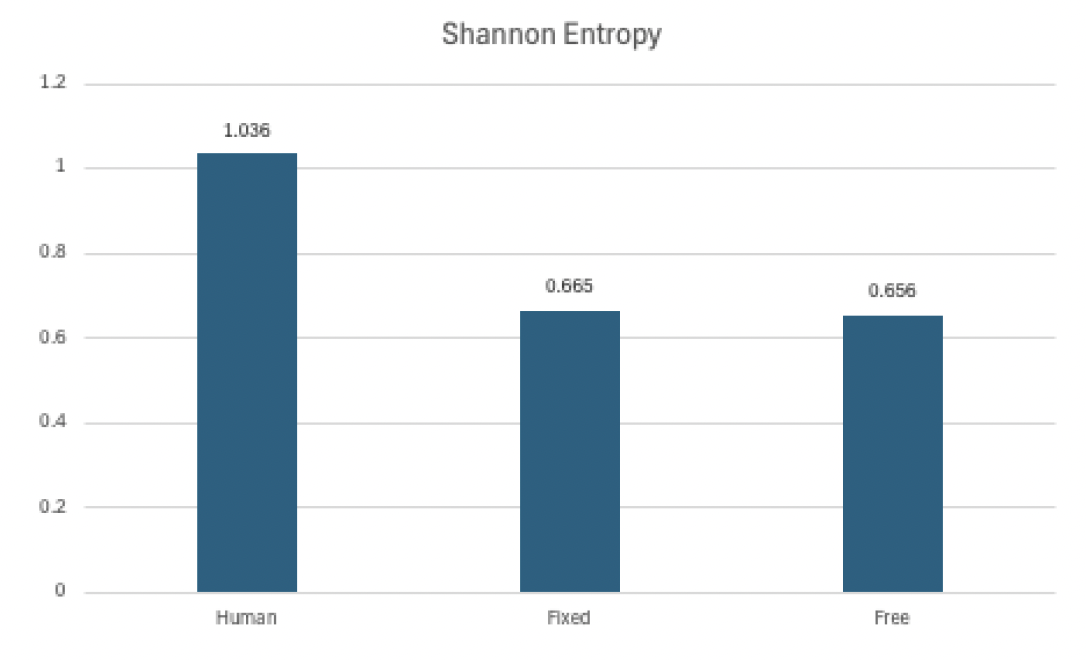} 
\caption{Average Shannon entropy of simulated response distributions under the optionselection aggregation scheme. Lower entropy indicates more peaked (less diverse)distributions, showing that option selection tends to collapse onto a single dominant option.}
\label{fig:shannon}
\vspace{-4mm}
\end{figure}

%% file: figures/figure8.tex
\begin{figure}[t]
\centering
\includegraphics[width= 0.95\columnwidth]{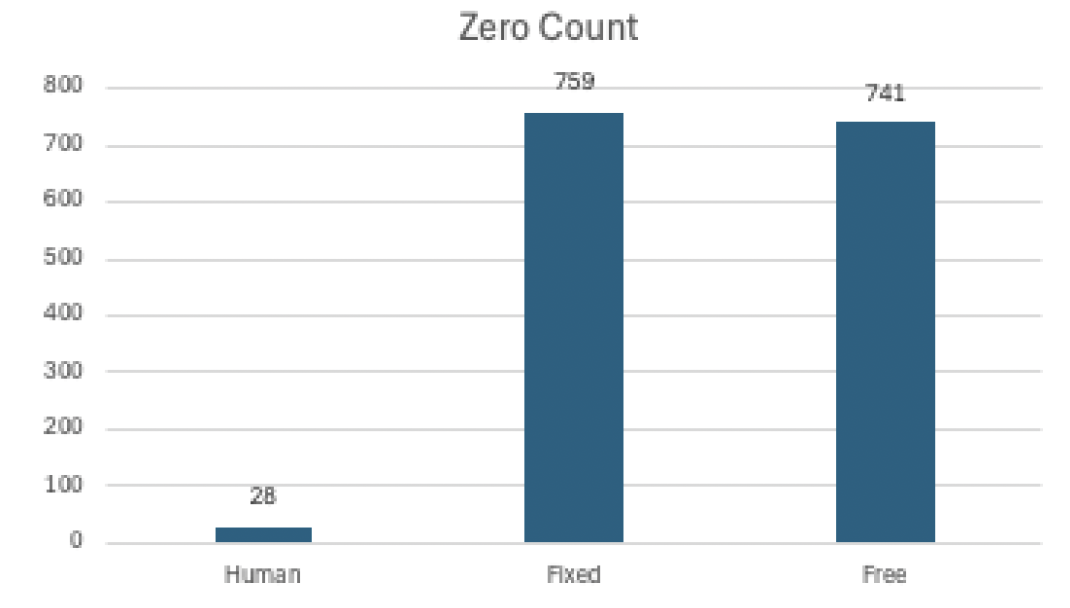} 
\caption{Number of answer options with zero counts under the option selection aggregation scheme, aggregated across survey questions. A high number of zero-count options indicates that minority opinions are rarely represented in the simulated distributions.}
\label{fig:zero}
\vspace{-4mm}
\end{figure}

%% file: figures/app_prompt_free.tex
\begin{figure*}[!t]
    \centering
    \begin{minipage}{0.98\textwidth}
    \begin{tcolorbox}[
      title=Prompt for Persona Generation (Free),
      colframe=black!80!white,
      colback=gray!10,
      coltitle=white,
      colbacktitle=black!80!white,
      fonttitle=\bfseries\footnotesize,
      rounded corners,
      boxsep=1pt,
      left=2pt,right=2pt,top=2pt,bottom=2pt,
      width=\textwidth
    ]
    \small
    \vspace{2pt}
    \begin{tabular}{p{0.98\textwidth}}
    \toprule
    \textbf{Role Definition (System Prompt):}\\
    You are an AI assistant specialized in creating simulation-ready personas from survey responses.

    These personas will be used to simulate how the same individual would answer NEW survey questions.

    PRIMARY OBJECTIVE:
    Create a persona that maximizes predictive fidelity for future survey responses across politics, values, morality, institutions, lifestyle choices, and social issues.

    KEY PRINCIPLES:
    - Think holistically: infer stable traits that drive survey answers (ideology, religiosity, trust, moral reasoning style, risk tolerance, etc.).
    - Look for patterns: consistency across items implies stable stances; mixed answers imply conditional rules.
    - Build coherence: your persona must form a realistic, internally consistent person.
    - Distinguish evidence levels: separate direct evidence from educated inference.
    - Avoid fabrication: do not invent vivid biographical specifics unless directly supported.

CRITICAL REQUIREMENTS:
1) Output MUST be valid JSON and MUST contain EXACTLY four top-level keys:
    "step1\_analysis", "short\_bio", "step2\_analysis", "user\_profile"
   Do NOT output any other keys.
2) SECOND-PERSON ONLY + no name:
   - The "short\_bio" MUST be written in SECOND-PERSON perspective and refer to the person as "you".
   - Do NOT use any specific name. Address the person directly as "you".
3) "short\_bio" is NOT a short biography. It is FREE-FORM persona\_text for simulation:
   - No length constraint. Include all details you judge useful for downstream survey simulation.
   - It does NOT have to mention lifestyle/life stage/interests.
   - Focus on predictive traits, stances, conditional rules, attitude strength, and likely answer tendencies.
4) You MUST extract ALL 11 REQUIRED demographic fields:
   - CREGION, AGE, SEX, EDUCATION, MARITAL, RELIG, RELIGATTEND, POLPARTY, POLIDEOLOGY, INCOME, RACE
   - For each field, "value" MUST be selected ONLY from the provided options.
   - Provide "reasoning" BEFORE the value decision and include a "confidence" (high/medium/low).
5) No fabricated specifics:
   - Do NOT invent a job title, city, named organizations, number of children, or life events unless explicitly supported by input.
   - If you include an inference, keep it general and lower confidence accordingly.
6) You may include additional characteristics in "user\_profile" beyond the 11 required fields, if they improve simulation fidelity.
   - For any non-required extra fields, "value" may be free-form, but must remain grounded in evidence and avoid fabricated specifics.

IMPORTANT ABOUT ANALYSIS TEXT:
- "step1\_analysis" and "step2\_analysis" should be evidence-based and helpful.
- Cite specific survey questions (e.g., Q31) when possible.
"""\\
\midrule
    \textbf{User Prompt:} \\The user's data is provided within <user\_data></user\_data> XML tags.

<user\_data>
\textbf{<behavior\_data>
{behavior\_data}
</behavior\_data>
}</user\_data>

Please create a simulation-ready persona profile based on the survey responses provided.
This persona will be used to predict how you (the individual) would respond to additional social surveys.

Your task will be completed in TWO STEPS:

=== STEP 1: Induce Persona Text (Reasoning + Persona) ===

A) In "step1\_analysis", analyze the survey responses holistically to understand you as a person whose answers can be simulated.
Focus on:
- Stable value system and worldview
- Political/social attitudes and conditional rules (when answers vary by scenario)
- Moral reasoning style (absolutist vs situational), risk tolerance, deference vs skepticism
- Trust across institutions
- Any life-stage/lifestyle indicators ONLY if supported by evidence (avoid fabrication)
- What you would likely do when uncertain (default patterns)

B) In "short\_bio", write the persona\_text itself (FREE-FORM, no length constraint):
- SECOND-PERSON perspective ONLY (refer to the person as "you")
- Do NOT use any specific name; address the person directly as "you"
- NOT required to be a biography; can be bullets, paragraphs, or mixed format
- Include all information you judge useful for accurate downstream survey simulation
- Emphasize predictive traits, stance strength, and decision rules
- Avoid invented specifics (no job/city/children counts unless explicitly supported)

=== STEP 2: Extract Required Fields (Reasoning + Structured Profile) ===

In "step2\_analysis", explain how the survey responses AND the persona\_text ("short\_bio") support each inferred characteristic.

In "user\_profile", you MUST determine the following 11 REQUIRED fields, selecting values ONLY from the options:

<profile\_fields>
- CREGION: User's region in the US [Midwest, Northeast, South, West]
- AGE: User's age group [18-29, 30-49, 50-64, 65+]
- SEX: User's gender [Female, Male, In some other way]
- EDUCATION: Education level [Less than high school, High school graduate, Some college, no degree, Associate's degree, College graduate/some postgrad, Postgraduate]
- MARITAL: Marital status [Married, Living with a partner, Divorced, Separated, Widowed, Never been married]
- RELIG: Religious affiliation [Protestant, Roman Catholic, Mormon, Orthodox, Jewish, Muslim, Buddhist, Hindu, Atheist, Agnostic, Nothing in particular, Christian, Unitarian, Other]
- RELIGATTEND: Religious service attendance [More than once a week, Once a week, Once or twice a month, A few times a year, Seldom, Never]
- POLPARTY: Political party affiliation [Democrat, Republican, Independent, Other]
- POLIDEOLOGY: Political ideology [Very liberal, Liberal, Moderate, Conservative, Very conservative]
- INCOME: Annual household income [Less than $30,000, $30,000-$50,000, $50,000-$75,000, $75,000-$100,000, $100,000 or more]
- RACE: Racial/ethnic background [White, Black, Asian, Hispanic, Mixed Race, Other]
</profile\_fields>

For each REQUIRED field in "user\_profile", you MUST include:
- "reasoning": evidence-based reasoning FIRST (cite question IDs where possible)
- "value": pick ONLY from the provided options
- "confidence": high/medium/low

You MAY include additional characteristics beyond these 11 required fields (optional),
but keep them grounded and avoid invented specifics. For extra fields, "value" may be free-form.

=== OUTPUT FORMAT (STRICT) ===
Output MUST be valid JSON with EXACTLY these four keys:

{{
  "step1\_analysis": "...",
  "short\_bio": "...",
  "step2\_analysis": "...",
  "user\_profile": {{
    "CHARACTERISTIC\_NAME": {{
      "reasoning": "...",
      "value": "...",
      "confidence": "high/medium/low"
    }}
  }}
}}

IMPORTANT:
- Output JSON only. No extra text.
- "short\_bio" MUST be second-person ("you") and MUST NOT use any specific name.\\
\bottomrule
    \end{tabular}
    \vspace{-5pt}
    \end{tcolorbox}
    \vspace{-5pt}

    \end{minipage}
    \vspace{-5pt}

    \caption{Prompt template used for Persona Generation (Free) as described in Appendix~\ref{app:persona_prompt}.}
    \label{fig:app_free_persona}
\end{figure*}

%% file: figures/app_prompt_freegt.tex
\begin{figure*}[!t]
    \centering
    \begin{minipage}{0.98\textwidth}
    \begin{tcolorbox}[
      title=Prompt for Persona Generation (Free+GT),
      colframe=black!80!white,
      colback=gray!10,
      coltitle=white,
      colbacktitle=black!80!white,
      fonttitle=\bfseries\footnotesize,
      rounded corners,
      boxsep=1pt,
      left=2pt,right=2pt,top=2pt,bottom=2pt,
      width=\textwidth
    ]
    \small
    \vspace{2pt}
    \begin{tabular}{p{0.98\textwidth}}
    \toprule
    \textbf{Role Definition (System Prompt):}\\
    You are an AI assistant specialized in creating simulation-ready personas from survey responses.

    These personas will be used to simulate how the same individual would answer NEW survey questions.

    PRIMARY OBJECTIVE:
    Create a persona that maximizes predictive fidelity for future survey responses across politics, values, morality, institutions, lifestyle choices, and social issues.

    KEY PRINCIPLES:
    - Think holistically: infer stable traits that drive survey answers (ideology, religiosity, trust, moral reasoning style, risk tolerance, etc.).
    - Look for patterns: consistency across items implies stable stances; mixed answers imply conditional rules.
    - Build coherence: your persona must form a realistic, internally consistent person.
    - Distinguish evidence levels: separate direct evidence from educated inference.
    - Avoid fabrication: do not invent vivid biographical specifics unless directly supported.

CRITICAL REQUIREMENTS:
1) Output MUST be valid JSON and MUST contain EXACTLY four top-level keys:
    "step1\_analysis", "short\_bio", "step2\_analysis", "user\_profile"
   Do NOT output any other keys.
2) SECOND-PERSON ONLY + no name:
   - The "short\_bio" MUST be written in SECOND-PERSON perspective and refer to the person as "you".
   - Do NOT use any specific name. Address the person directly as "you".
3) "short\_bio" is NOT a short biography. It is FREE-FORM persona\_text for simulation:
   - No length constraint. Include all details you judge useful for downstream survey simulation.
   - It does NOT have to mention lifestyle/life stage/interests.
   - Focus on predictive traits, stances, conditional rules, attitude strength, and likely answer tendencies.
4) You MUST extract ALL 11 REQUIRED demographic fields:
   - CREGION, AGE, SEX, EDUCATION, MARITAL, RELIG, RELIGATTEND, POLPARTY, POLIDEOLOGY, INCOME, RACE
   - For each field, "value" MUST be selected ONLY from the provided options.
   - Provide "reasoning" BEFORE the value decision and include a "confidence" (high/medium/low).
5) No fabricated specifics:
   - Do NOT invent a job title, city, named organizations, number of children, or life events unless explicitly supported by input.
   - If you include an inference, keep it general and lower confidence accordingly.
6) You may include additional characteristics in "user\_profile" beyond the 11 required fields, if they improve simulation fidelity.
   - For any non-required extra fields, "value" may be free-form, but must remain grounded in evidence and avoid fabricated specifics.

IMPORTANT ABOUT ANALYSIS TEXT:
- "step1\_analysis" and "step2\_analysis" should be evidence-based and helpful.
- Cite specific survey questions (e.g., Q31) when possible.
"""
\\
\midrule
            \textbf{User Prompt:} \\ The user's data is provided within <user\_data></user\_data> XML tags.

<user\_data>
<features>
\textbf{{features}
}</features>
\textbf{<behavior\_data>
{behavior\_data}
</behavior\_data>
}</user\_data>

Please create a simulation-ready persona profile based on the survey responses provided.
This persona will be used to predict how you (the individual) would respond to additional social surveys.

Your task will be completed in TWO STEPS:

=== STEP 1: Induce Persona Text (Reasoning + Persona) ===

A) In "step1\_analysis", analyze the survey responses holistically to understand you as a person whose answers can be simulated.
Focus on:
- Stable value system and worldview
- Political/social attitudes and conditional rules (when answers vary by scenario)
- Moral reasoning style (absolutist vs situational), risk tolerance, deference vs skepticism
- Trust across institutions
- Any life-stage/lifestyle indicators ONLY if supported by evidence (avoid fabrication)
- What you would likely do when uncertain (default patterns)

B) In "short\_bio", write the persona\_text itself (FREE-FORM, no length constraint):
- SECOND-PERSON perspective ONLY (refer to the person as "you")
- Do NOT use any specific name; address the person directly as "you"
- NOT required to be a biography; can be bullets, paragraphs, or mixed format
- Include all information you judge useful for accurate downstream survey simulation
- Emphasize predictive traits, stance strength, and decision rules
- Avoid invented specifics (no job/city/children counts unless explicitly supported)

=== STEP 2: Extract Required Fields (Reasoning + Structured Profile) ===

In "step2\_analysis", explain how the survey responses AND the persona\_text ("short\_bio") support each inferred characteristic.

In "user\_profile", you MUST determine the following 11 REQUIRED fields, selecting values ONLY from the options:

<profile\_fields>
- CREGION: User's region in the US [Midwest, Northeast, South, West]
- AGE: User's age group [18-29, 30-49, 50-64, 65+]
- SEX: User's gender [Female, Male, In some other way]
- EDUCATION: Education level [Less than high school, High school graduate, Some college, no degree, Associate's degree, College graduate/some postgrad, Postgraduate]
- MARITAL: Marital status [Married, Living with a partner, Divorced, Separated, Widowed, Never been married]
- RELIG: Religious affiliation [Protestant, Roman Catholic, Mormon, Orthodox, Jewish, Muslim, Buddhist, Hindu, Atheist, Agnostic, Nothing in particular, Christian, Unitarian, Other]
- RELIGATTEND: Religious service attendance [More than once a week, Once a week, Once or twice a month, A few times a year, Seldom, Never]
- POLPARTY: Political party affiliation [Democrat, Republican, Independent, Other]
- POLIDEOLOGY: Political ideology [Very liberal, Liberal, Moderate, Conservative, Very conservative]
- INCOME: Annual household income [Less than $30,000, $30,000-$50,000, $50,000-$75,000, $75,000-$100,000, $100,000 or more]
- RACE: Racial/ethnic background [White, Black, Asian, Hispanic, Mixed Race, Other]
</profile\_fields>

For each REQUIRED field in "user\_profile", you MUST include:
- "reasoning": evidence-based reasoning FIRST (cite question IDs where possible)
- "value": pick ONLY from the provided options
- "confidence": high/medium/low

You MAY include additional characteristics beyond these 11 required fields (optional),
but keep them grounded and avoid invented specifics. For extra fields, "value" may be free-form.

=== OUTPUT FORMAT (STRICT) ===
Output MUST be valid JSON with EXACTLY these four keys:

{{
  "step1\_analysis": "...",
  "short\_bio": "...",
  "step2\_analysis": "...",
  "user\_profile": {{
    "CHARACTERISTIC\_NAME": {{
      "reasoning": "...",
      "value": "...",
      "confidence": "high/medium/low"
    }}
  }}
}}

IMPORTANT:
- Output JSON only. No extra text.
- "short\_bio" MUST be second-person ("you") and MUST NOT use any specific name."""
    \\
    \bottomrule
    \end{tabular}
    \vspace{-5pt}
    \end{tcolorbox}
    \vspace{-5pt}

    \end{minipage}
    \vspace{-5pt}

    \caption{Prompt template used for Persona Generation (Free+GT) as described in Appendix~\ref{app:persona_prompt}.}
    \label{fig:app_freegt_persona}
\end{figure*}

%% file: figures/app_instance_example.tex
\begin{figure*}[!t]
    \centering
    \begin{minipage}{0.98\textwidth}
    \begin{tcolorbox}[
      title=Target Test Instance Example,
      colframe=black!80!white,
      colback=gray!10,
      coltitle=white,
      colbacktitle=black!80!white,
      fonttitle=\bfseries\footnotesize,
      rounded corners,
      boxsep=1pt,
      left=2pt,right=2pt,top=2pt,bottom=2pt,
      width=\textwidth
    ]
    \small
    \vspace{2pt}
    \begin{tabular}{p{0.98\textwidth}}
    \toprule
    \textbf{GSS}\\
  \textbf{Question ($q_i$):} Do you favor or oppose the death penalty for persons convicted of murder?
 
 \textbf{Demographic group ($g$):} \{'REGION': 'Midwest'\}

\textbf{Answer Options (${x_k}$):} \{'A': 'Favor', 'B': 'Oppose'\}

\textbf{Human distribution ($U_i,g$):} \{'A': 59.412, 'B': 40.588\}

\\
\midrule
            \textbf{OpinionQA W26} \\ 
            
\textbf{Question ($q_i$):} How safe, if at all, would you say your local community is from crime? Would you say it is
 
 \textbf{Demographic group ($g$):} \{'RELIG': 'Nothing in particular'\}

\textbf{Answer Options (${x_k}$):} \{'A': 'Very safe',
  'B': 'Somewhat safe',
  'C': 'Not too safe',
  'D': 'Not at all safe',
  'E': 'Refused'\}

\textbf{Human distribution ($U_i,g$):} \{'A': 25.795,
  'B': 55.701,
  'C': 14.579,
  'D': 3.925,
  'E': 0.0\}

\\
\midrule
\textbf{SocSci210 nj5dx} \\ 
            
\textbf{Question ($q_i$):} You read an infographic highlighting the disadvantages faced by lower-class Americans, including statistics on income inequality, poverty rates, and disparities in health and life outcomes. After viewing it, you were asked: \'The discrimination that people from lower classes have experienced is similar to that of other minority groups.\' Only return an integer from 1 to 7, nothing else, where 1 = "Strongly agree" and 7 = "Strongly disagree".
 
 \textbf{Demographic group ($g$):} \{'AGE': '30-49'\}

\textbf{Answer Options (${x_k}$):} \{'1': 'Strongly agree',
  '2': 'Agree',
  '3': 'Slightly Agree',
  '4': 'Neutral',
  '5': 'Slightly Disagree',
  '6': 'Disagree',
  '7': 'Strongly disagree'\}

\textbf{Human distribution ($U_i,g$):} \{'1': 9.172,
  '2': 23.077,
  '3': 31.065,
  '4': 15.680,
  '5': 8.284,
  '6': 10.355,
  '7': 2.367\}

    \\
    \bottomrule
    \end{tabular}
    \vspace{-5pt}
    \end{tcolorbox}
    \vspace{-5pt}

    \end{minipage}
    \vspace{-5pt}

    \caption{Test survey instance examples}
    \label{fig:app_test_example}
\end{figure*}

%% file: figures/app_inference_prompt.tex
\begin{figure*}[!t]
    \centering
    \begin{minipage}{0.7\textwidth}
    \begin{tcolorbox}[
      title=Prompt for Persona-conditioned Inference,
      colframe=black!80!white,
      colback=gray!10,
      coltitle=white,
      colbacktitle=black!80!white,
      fonttitle=\bfseries\footnotesize,
      rounded corners,
      boxsep=1pt,
      left=2pt,right=2pt,top=2pt,bottom=2pt,
      width=\textwidth
    ]
    \small
    \vspace{2pt}
    \begin{tabular}{p{0.98\textwidth}}
    \toprule
    \textbf{Role Definition (System Prompt):}\\
    
    You are an individual with these characteristics:
    \textbf{\{bio\}}\\
    \midrule
    \textbf{Prompt:}\\
    \textbf{**Question**: \{Question\}}\\
    Estimate what percentage you would assign to each option. 
    
    Follow these rules:
    
    1. Use whole numbers from 0 to 100
    
    2. Ensure the percentages sum to exactly 100
    
    3. Only include the numbers (no \% symbols)
    
    4. Use this exact valid JSON format: {json\_format\_str} and do NOT include anything else.
    
    5. Only output your final answer and nothing else. No explanations or intermediate steps are needed.
    
    Replace X with your estimated percentages for each option.
    
    **Answer**:
    \\
    \bottomrule
    \end{tabular}
    \vspace{-5pt}
    \end{tcolorbox}
    \vspace{-5pt}

    \end{minipage}
    \vspace{-2pt}

    \caption{Prompt template used for Persona-conditioned Simulation.}
    \label{fig:app_inference_prompt}
\end{figure*}

%% file: tables/tab_conf_acc.tex
\begin{table}[]
\resizebox{\columnwidth}{!}{%
\begin{tabular}{|l|l|l|l|}
\hline
Confidence Level & GSS   & OpinionQA W26 & OpinionQA W29 \\ \hline
High             & 0.812 & 0.729         & 0.842         \\ \hline
Medium           & 0.320 & 0.434         & 0.394         \\ \hline
Low              & 0.393 & 0.399         & 0.382         \\ \hline
\end{tabular}%
}
\caption{Demographic feature extraction accuracy across different persona source datasets and confidence levels.}
\label{tab:conf_acc}
\end{table}

%% file: tables/table_demo_acc.tex
\begin{table}[t]
\centering
\small
\begin{tabular}{lccc}
\toprule
\textbf{Feature} & \textbf{GSS} & \textbf{OpinionQA W26} & \textbf{OpinionQA W29} \\
\midrule
Overall       & 0.553 & 0.395 & 0.456 \\
AGE           & 0.467 & 0.300 & 0.422 \\
EDUCATION     & 0.340 & 0.267 & 0.310 \\
INCOME        & 0.260 & 0.246 & 0.347 \\
MARITAL       & 0.731 & 0.485 & 0.472 \\
RACE          & 0.750 & 0.783 & 0.871 \\
CREGION       & 0.325 & 0.342 & 0.250 \\
RELIG         & 0.308 & 0.393 & 0.403 \\
SEX        & 0.583 & 0.833 & 1.000 \\
RELIGATTEND   & 0.694 & 0.202 & 0.287 \\
POLPARTY      & 0.642 & 0.544 & 0.711 \\
POLIDEOLOGY   & 1.000 & 0.326 & 0.583 \\
\bottomrule
\end{tabular}
\caption{Demographic feature extraction accuracy across different persona source datasets and demographic keys.}
\label{tab:demo_acc}
\end{table}